\documentclass{article}

\usepackage{PRIMEarxiv}

\usepackage{amsmath,amsfonts,bm}

\def\eqref#1{equation~\ref{#1}}

\def\1{\bm{1}}

\DeclareMathAlphabet{\mathsfit}{\encodingdefault}{\sfdefault}{m}{sl}
\SetMathAlphabet{\mathsfit}{bold}{\encodingdefault}{\sfdefault}{bx}{n}

\usepackage{hyperref}
\usepackage{url}
\usepackage{graphicx}
\usepackage{adjustbox}
\usepackage{booktabs}
\usepackage{natbib}

\usepackage{soul}
\usepackage[textsize=tiny, textwidth=1.23cm]{todonotes}

\usepackage[utf8]{inputenc} 
\usepackage[T1]{fontenc}    
\usepackage{hyperref}       
\usepackage{url}            
\usepackage{booktabs}       
\usepackage{amsfonts}       
\usepackage{nicefrac}       
\usepackage{microtype}      
\usepackage{xcolor}         
\usepackage{amsmath}
\newcommand{\itap}{I-TAP}
\usepackage{hyperref}
\usepackage{url}
\usepackage{tabularx}
\usepackage{graphicx}
\usepackage{multirow}
\usepackage{array}
\usepackage{siunitx}
\usepackage{adjustbox}
\usepackage{caption}
\usepackage{wrapfig}
\usepackage{float}
\usepackage{subcaption}
\usepackage{graphicx}
\usepackage{titlesec}
\usepackage{algorithm}
\usepackage{algpseudocode}
\usepackage{amssymb} 
\usepackage{cleveref}
\usepackage{xcolor}
\usepackage[normalem]{ulem}
\usepackage{tikz}
\usepackage{mathtools,graphicx}
\usepackage{natbib}
\usetikzlibrary{arrows.meta,positioning,calc,fit,backgrounds,matrix,decorations.pathreplacing}

\makeatletter
\newenvironment{breakablealgorithm}
  {%
   \begin{center}
   \refstepcounter{algorithm}%
   \captionsetup{type=algorithm}
   \hrule height .8pt depth 0pt \kern 2pt
   \renewcommand{\caption}[2][\relax]{%
      {\raggedright\textbf{\fname@algorithm~\thealgorithm}\quad ##2\par}%
      \ifx\relax##1\relax\else{\raggedright\footnotesize ##1\par}\fi
      \kern2pt\hrule\kern2pt
   }%
  }{%
   \kern2pt\hrule
   \end{center}
  }
\makeatother

\newcommand{\pastK}{k_{t_{k-1}:t_{k-c},\,1:D}}
\newcommand{\Ck}{\mathcal{C}_k}

\newcommand{\Emb}[1]{\mathrm{Embed}\!\big(#1\big)}
\newcommand{\TopKTempCat}{\textsc{TopKTempCat}}

\newcommand{\RTG}[1]{\big[#1\big]_{\mathrm{rtg}}}            
\newcommand{\SampleStack}{\operatorname{SampleStack}}
\newcommand{\Obs}{\operatorname{Obs}}

\title{In‑Context Planning with Latent Temporal Abstractions}

\author{
\textbf{Baiting Luo} \quad
\textbf{Yunuo Zhang} \quad
\textbf{Nathaniel S.~Keplinger} \quad
\textbf{Samir Gupta} \quad
\textbf{Abhishek Dubey} \\
Vanderbilt University \\
\texttt{\{baiting.luo,yunuo.zhang,nathaniel.s.keplinger,} \\
\texttt{samir.gupta,abhishek.dubey\}@vanderbilt.edu} \\[8pt]
\textbf{Ayan Mukhopadhyay} \\
William \& Mary \\
\texttt{ayan.mukhopadhyay@wm.edu}
}
\begin{document}

\maketitle


\begin{abstract}
Planning-based reinforcement learning for continuous control is bottlenecked by two practical issues: planning at primitive time scales leads to prohibitive branching and long horizons, while real environments are frequently partially observable and exhibit regime shifts that invalidate stationary, fully observed dynamics assumptions. We introduce I‑TAP (In‑Context Latent Temporal‑Abstraction Planner), an offline RL framework that unifies in-context adaptation with online planning in a learned discrete temporal-abstraction space. From offline trajectories, I‑TAP learns an observation-conditioned residual-quantization VAE that compresses each observation–macro-action segment into a coarse-to-fine stack of discrete residual tokens, and a temporal Transformer that autoregressively predicts these token stacks from a short recent history. The resulting sequence model acts simultaneously as a context-conditioned prior over abstract actions and a latent dynamics model. At test time, I‑TAP performs Monte Carlo Tree Search directly in token space, using short histories for implicit adaptation without gradient update, and decodes selected token stacks into executable actions. Across deterministic MuJoCo, stochastic MuJoCo with per episode latent dynamics regimes, and high-dimensional Adroit manipulation, including partially observable variants, I-TAP consistently matches or outperforms strong model-free and model-based offline baselines, demonstrating efficient and robust in-context planning under stochastic dynamics and partial observability.
\end{abstract}

\section{Introduction}

\begingroup
\renewcommand{\thefootnote}{}
\footnotetext{Source code: \url{https://github.com/BaitingLuo/I-TAP}}
\endgroup

Planning-based reinforcement learning (RL) has delivered strong results in discrete decision-making domains (e.g., board games and video games)~\citep{silver2017masteringchessshogiselfplay, DBLP:journals/nature/SchrittwieserAH20} and has shown increasing promise in continuous control~\citep{DBLP:conf/icml/HubertSABSS21}. However, planning-based offline RL for continuous control faces two practical challenges. First, many real environments are effectively partially observable due to latent parameters (e.g., unobserved disturbances or payload changes), which breaks the stationary, full observability assumptions commonly adopted by learned dynamics models used for planning. When these latent factors are not properly handled during planning, they manifest as apparent stochasticity from the planner's perspective~\citep{DBLP:conf/iclr/AntonoglouSOHS22}, exacerbating planning complexity. Second, planning over primitive continuous actions induces large branching factors and long effective horizons, making search expensive especially under uncertainty.

To address adaptation under partial observability, recent work has reframed RL as sequence modeling, leveraging Transformers~\citep{DBLP:conf/nips/VaswaniSPUJGKP17} trained on trajectories to produce policies conditioned on a finite context window~\citep{DBLP:conf/nips/ChenLRLGLASM21, DBLP:conf/nips/BrownMRSKDNSSAA20,furuta2022generalized,DBLP:conf/icml/LiuA23,DBLP:conf/iclr/LaskinWOPSSSHFB23,DBLP:conf/icml/HuangHCS024}. Conditioning on recent interaction history enables in-context adaptation without test-time gradient updates, and can implicitly capture latent task variables when they are identifiable from history. Yet these sequence-model policies are typically deployed as direct action predictors, which introduces two limitations: (i) Without an explicit decision-time optimizer, they often inherit the constraints and suboptimalities of the offline dataset~\citep{son2025distilling}. (ii) In stochastic environments, converting a predictive model into an optimized decision rule is nontrivial when the model is used only as a conditional policy~\citep{DBLP:conf/nips/PasterMB22}. These issues motivate combining in-context models with explicit planning.

Nevertheless, planning directly in raw continuous action space can be inefficient and inflexible~\citep{DBLP:conf/iclr/JiangZJLRGT23}. Recent planning-based methods reduce complexity by learning temporal abstractions (e.g., options~\citep{SUTTON1999181} or macro-actions~\citep{macro-action}) and planning over high-level decisions~\citep{DBLP:conf/iclr/JiangZJLRGT23,luo2025scalable}, which shortens effective horizons and reduces branching. However, these planners are developed under the assumption of a stationary, fully observed Markov decision process (MDP) and learn dynamics models that condition only on the current state. As a result, state aliasing would create apparent stochasticity and increase the search burden.

Motivated by these challenges, we introduce an in-context planning framework that adapts to different latent parameters in a learned discrete latent temporal abstraction space for continuous control under stochastic dynamics. Our premise is that integrating temporal abstraction with in-context adaptation for planning-based RL addresses these issues jointly. Adapting from recent history and planning conditioned on it in a latent temporal abstraction space allows an agent to: (i) decouple adaptation and planning from the native temporal granularity of MDP, thereby shortening the required context, easing the learning of a reliable sequence model prior by modeling a simpler distribution over discrete latent temporal abstractions instead of high-dimensional continuous actions, and reducing the branching factor during planning; (ii) use context to infer global latent parameters that govern the environment’s dynamics (e.g., unobserved perturbation forces), enabling effective adaptation and forecasting across scenario shifts; and (iii) employ an online planner such as Monte Carlo Tree Search (MCTS) for optimization, providing a mechanism to deviate from suboptimal behavior policies in the offline data and handle uncertainty.

\begin{figure*}
\includegraphics[width=\columnwidth]{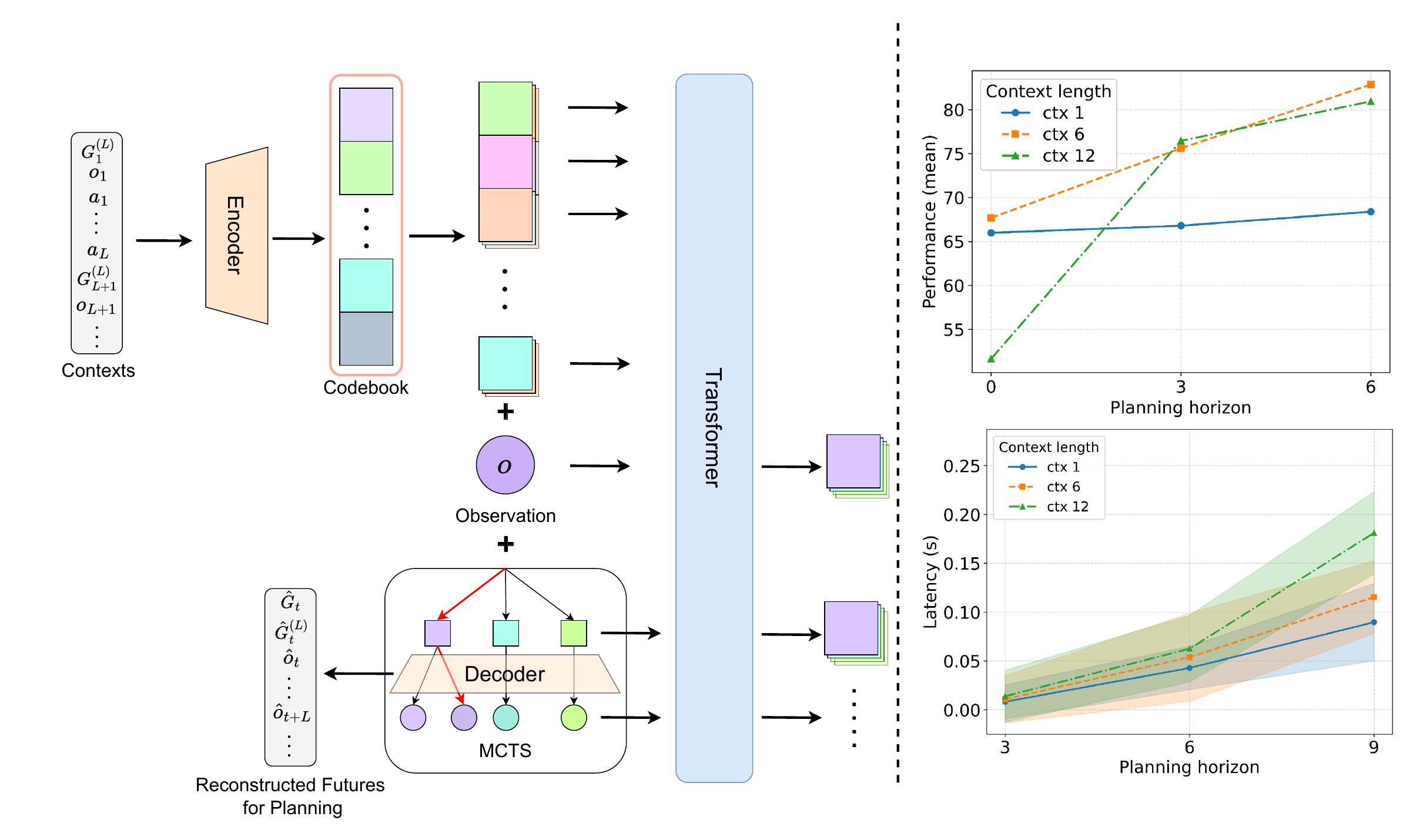}
\caption{Overview of \itap. \textbf{Left:} A residual-quantized VAE (RQ-VAE) discretizes continuous observation–action trajectories into a coarse-to-fine token stack. \textbf{Right:} Normalized return and per-decision latency as functions of planning horizon and context size on Stochastic MuJoCo, highlighting the importance of a properly sized context window for effective in-context planning under environmental stochasticity and partial observability.}
\label{fig:overview}
\end{figure*}

To this end, we propose \textit{In-Context Latent Temporal Abstraction Planner} (\itap), which learns a discrete latent temporal-abstraction space and a Transformer-based in-context sequence prior over discrete latent codes from offline data, and performs decision-time planning with these models to enable adaptive decision-making. To improve flexibility and scalability in handling high-dimensional continuous observation-action spaces, we adopt an observation-conditioned residual-quantized VAE (RQ-VAE) to learn this discrete latent temporal-abstraction space, as illustrated in Fig.~\ref{fig:overview}. RQ-VAE~\citep{DBLP:conf/cvpr/LeeKKCH22} encodes each macro-step into a depth-$D$ coarse-to-fine stack of code indices drawn from a codebook, providing a compositional discretization with substantially higher effective capacity than a single code. This retains high-fidelity decoding while enabling compact discrete representations, yielding a scalable interface for in-context sequence modeling over abstractions and planning in latent space. MCTS then operates over these latent tokens using context\mbox{-}guided priors to balance exploration and exploitation under uncertainty; finally, we decode the selected latent stack to a primitive action sequence and execute the first action. 

Our experiments demonstrate that \itap\ can be trained as a single offline model across behavior policies of varying quality and multiple latent parameters, and evaluated from deterministic to highly stochastic environments. Across these settings, \itap\ matches or outperforms strong offline RL and planning-based baselines, while exhibiting in-context adaptation under partial observability and scaling to high-dimensional continuous control through latent temporal abstractions. In summary, we make the following contributions:
(i) We propose \itap, an offline RL method that unifies in-context adaptation and online planning for stochastic, partially observable continuous control by planning in a learned latent temporal-abstraction space;
(ii) To improve scalability and flexibility, we learn a residual-quantized temporal abstraction and a context-conditioned latent dynamics model from offline trajectories, via an observation-conditioned RQ-VAE and an autoregressive temporal Transformer;
(iii) We instantiate an in-context planner that performs Monte Carlo Tree Search directly over latent tokens, reducing the effective branching factor and decision horizon, and enabling improvements beyond suboptimal behavior policies.

\section{Problem Formulation}

\paragraph{POMDP.}
We consider a partially observable Markov decision process (POMDP)
$M=(\mathcal{S},\mathcal{A},\mathcal{O},P,R,\gamma)$, where
$P(s_{t+1}\mid s_t,a_t)$ is the transition kernel, $R(r_t\mid s_t,a_t)$ is the reward distribution,
and $\gamma\in[0,1)$ is the discount factor.
At each time step $t$, the agent receives an observation $o_t\in\mathcal{O}$, selects an action $a_t\in\mathcal{A}$,
the environment transitions according to $P(s_{t+1}\mid s_t,a_t)$, and the agent receives a reward
$r_t\sim R(\cdot\mid s_t,a_t)$. For a fixed context length $c$, we define the agent's context prior to observing $o_t$ as
$c_t=(o_{t-c},a_{t-c},r_{t-c},\ldots,o_{t-1},a_{t-1},r_{t-1})\in\mathcal{C}$. The goal is to maximize the expected discounted return
$J=\mathbb{E}_{\pi,M}\!\left[\sum_{t=0}^{T-1}\gamma^t r_t\right]$ over horizon $T$.


\paragraph{Meta-RL.}
We define a context-conditioned policy
$\pi:\mathcal{C}\times\mathcal{O}\rightarrow \Delta(\mathcal{A})$,
where $\mathcal{C}$ is a fixed-length context space and $\Delta(\mathcal{A})$ denotes distributions over actions.
At time $t$, the algorithm forms a bounded context $c_t\in\mathcal{C}$ from the interaction history $h_t$
(e.g., a sliding window of recent transitions) and samples actions as $a_t\sim f(c_t,o_t)$. Meta-RL aims to learn an algorithm $\pi$ that maximizes $J$ in expectation over a distribution of tasks (POMDPs) $p(M)$:
$\max_\pi \ \mathbb{E}_{M\sim p(M)}\left[\mathbb{E}_{\pi,M}\!\left[\sum_{t=0}^{H-1}\gamma^t r_t\right]\right].$
In our setting, each $M$ is induced by an unobserved latent
parameter. In offline meta-RL, we assume access to a dataset of trajectories collected by some behavior algorithms on meta-training tasks, from which the context $c_t$ is constructed.

\section{Background}\label{sec:background}
\paragraph{Residual-Quantized Variational Autoencoder (RQ-VAE).}
Residual-Quantized VAE (RQ-VAE)~\citep{DBLP:conf/cvpr/LeeKKCH22} is a discrete autoencoding model that
replaces the single-step vector quantization in VQ-based models with residual quantization, representing
each latent as a stack of $D$ discrete codes. Let $z_t \in \mathbb{R}^d$ denote a continuous feature vector
produced by an encoder, and let the codebook be $E=\{e_1,\dots,e_K\}\subset\mathbb{R}^d$. RQ-VAE iteratively
quantizes residuals in a coarse-to-fine manner by initializing $r_t^{(0)} := z_t$ and, for depths $\ell=1,\dots,D$,
computing
\[
k_{t,\ell}=\arg\min_{k\in[K]}\|r_t^{(\ell-1)}-e_k\|_2^2,\qquad
r_t^{(\ell)} = r_t^{(\ell-1)} - e_{k_{t,\ell}},
\]
which yields the depth-$\ell$ partial sum $\hat z_t^{(\ell)} := \sum_{j=1}^{\ell} e_{k_{t,j}}$ and the final quantized
vector $\hat z_t := \hat z_t^{(D)}$. Using a shared codebook across depths yields an effective representational
capacity that grows as $K^D$ without enlarging $K$. The decoder reconstructs the input from $\hat z_t$, and the
model is trained jointly with minimizing a reconstruction loss plus a depth-wise commitment regularizer,
using the straight-through estimator for backpropagation through discrete assignments and updating the
codebook via exponential moving average (EMA)~\citep{DBLP:conf/nips/OordVK17}. An RQ-Transformer~\citep{DBLP:conf/cvpr/LeeKKCH22} combines a spatial Transformer that summarizes codes from previous positions into a context vector with a depth Transformer that, conditioned on this context, predicts the $D$ codes within the current stack in a coarse-to-fine manner, thereby modeling an autoregressive prior over code stacks.


\paragraph{Temporal Abstraction via Macro-Actions.}
We introduce temporal abstractions via a fixed macro length $L$ and a macro-action space
$\mathcal{A}^{(L)}$, where each macro-action $m_b\in\mathcal{A}^{(L)}$ consists of $L$ primitive actions. Decisions are made at macro-steps $b=0,1,\ldots,N-1$, corresponding to primitive times $t=bL$,
so the episode horizon satisfies $T = NL$.
At macro-step $b$, the agent selects $m_b$ and executes it for $L$ primitive steps, generating rewards
$\{r_{bL},\ldots,r_{bL+L-1}\}$.

\paragraph{Trajectory Representation.}
Consider an episode with $T=NL$ primitive steps and macro boundaries at $t=bL$.
The return-to-go from primitive time $t$ is $G_t = \sum_{i=t}^{T-1} \gamma^{\,i-t} r_i$,
and the $L$-step discounted return from a macro boundary is $G^{(L)}_{bL}=\sum_{i=0}^{L-1}\gamma^i r_{bL+i}$. We write the macro-level trajectory as $\tau = \bigl((G_{bL},\, G^{(L)}_{bL},\, o_{bL},\, m_b)\bigr)_{b=0}^{N-1}$, where $o_{bL}$ is the observation at the macro boundary and $m_b$ is the macro-action executed for
the subsequent $L$ primitive steps. This representation preserves the underlying primitive dynamics while exposing temporally extended actions.

\section{Method}


In this section, we provide an in-depth explanation of each component of {\itap}: the discretization of observation-macro-actions sequences, modeling the in-context prior and transition distribution over latent codes, and the planning process with MCTS. In general, {\itap} is a generative model for trajectories conditioned on both current observation and recent historical context, allowing for efficient in-context planning and decision-making.

\subsection{In-Context Residual Discretization of Observation Macro Action Sequences}
In-context RL approaches tend to replicate the suboptimal behaviors of the source algorithm~\citep{son2025distilling}, necessitating the integration of a planning mechanism to deviate from suboptimal decisions. Meanwhile, a discrete action space simplifies the representation of action distributions and facilitates the use of advanced planning algorithms~\citep{doi:10.1126/science.aar6404}.

To leverage these advantages, prior work has proposed state-conditioned Vector Quantized VAEs (VQ-VAEs)~\citep{DBLP:conf/iclr/JiangZJLRGT23}
to discretize continuous control into a compact latent action space. However, single-code vector quantization can become a bottleneck:
to maintain reconstruction fidelity for high-dimensional features, one often needs either a very large codebook (inflating the discrete vocabulary)
or additional compression, both of which can degrade reconstruction quality when features are high-dimensional~\citep{DBLP:conf/cvpr/LeeKKCH22,DBLP:conf/iclr/JiangXWLJGRT24}.
In our setting, each macro-step token must summarize an observation together with a temporally extended action, and the encoder processes these as context-dependent sequences, making the embeddings more context-dependent and diverse, which further increases the required quantization capacity.
We therefore adopt an observation-conditioned RQ-VAE, which represents each macro-step with a depth-$D$ coarse-to-fine stack of codes, providing substantially higher effective capacity
without requiring an excessively large single-step codebook. This yields compact discrete representations while preserving decoding fidelity, which is crucial when
downstream decision making depends on model predictions over these abstractions.

\begin{figure*}
    \centering
    \includegraphics[width=\columnwidth]{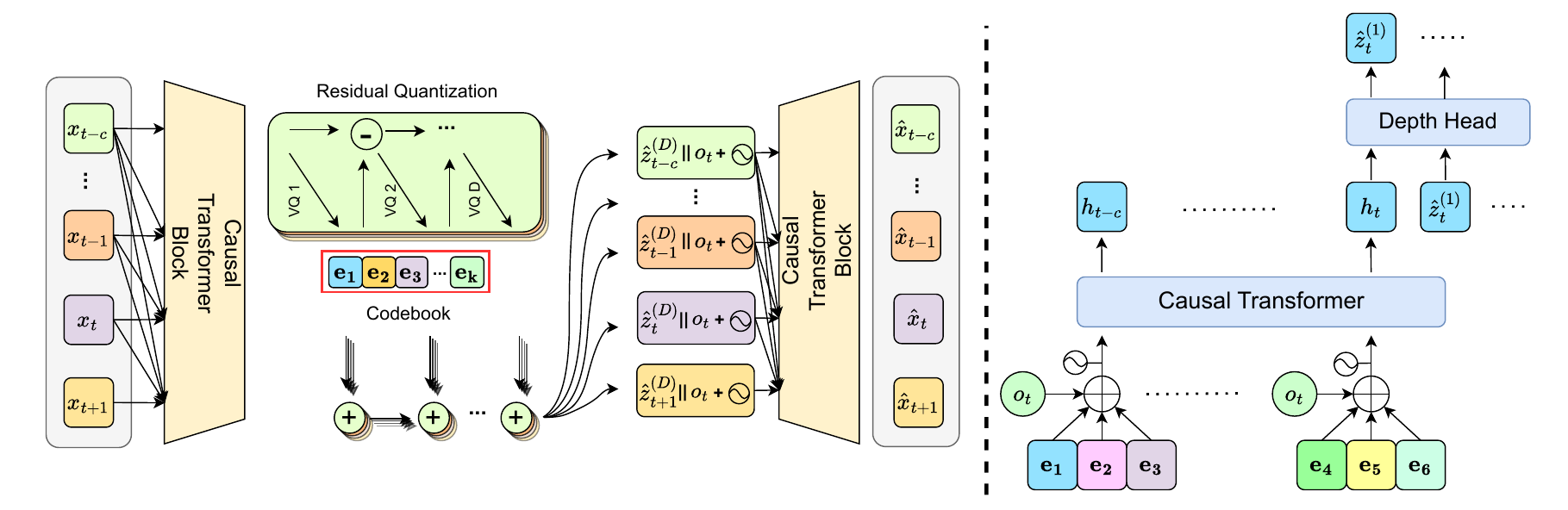}
    \caption{An overview of our RQ-VAE model that discretizes state-macro action sequences and temporal prior for {\itap}}
    \label{fig:model_training}
\end{figure*}

\paragraph{Tokens and masking.}
At macro index $b$, a token is
$x_{t_b}=\big(G_{t_b},\,G^{(L)}_{t_b},\,o_{t_b},\,m_{t_b}\big)$.
During training, we sample contiguous chunks $(x_{t_{b-c}},\,x_{t_{b-c+1}},\,\ldots,\,x_{t_b},\,x_{t_{b+1}})$,
where the first $c$ tokens provide context and the last two tokens correspond to the current and next macro-step.

Inspired by~\citet{luo2025scalable}, we mask return components to reduce variance induced by behavior-policy quality and environmental stochasticity.
Specifically, we mask the return-to-go $G_t$ at all positions, since it is a high-variance long-horizon signal and requires choosing a test-time target value.
We retain the short-horizon return $G^{(L)}_t$ only in the context tokens and mask it for the current and next tokens $(x_{t_b}, x_{t_{b+1}})$.
Consequently, the code assigned to the current macro decision is determined by the observation--macro-action pair and the preceding context, rather than by the realized return, which reduces sensitivity to return variability in the offline data.

This design avoids manually choosing a target $G_{t_b}$, which is susceptible to luck-induced variability in stochastic environments~\citep{DBLP:conf/nips/PasterMB22},
while still using $G^{(L)}$ in the context as a stable short-horizon signal for inferring latent regimes.





\paragraph{Discretization with RQ-VAE} A causal Transformer maps the chunk to per-token features $Z \;=\; \big(z_{t_{b-c}},\ldots,z_{t_b},z_{t_{b+1}}\big)$. For each token index $t \in \{t_b-c,\ldots,t_b+1\}$, we apply residual quantization (\cref{sec:background}) with a shared
codebook $E=\{e_1,\ldots,e_K\}\subset\mathbb{R}^d$ to obtain a depth-$D$ code stack $k_{t,1:D}$ and partial sums
$\hat z_t^{(\ell)}=\sum_{j=1}^\ell e_{k_{t,j}}$; we use the final quantized embedding $\hat z_t^{(D)}$ as the discrete
representation of $x_t$.

Motivated by evidence from prior work~\citep{DBLP:conf/iclr/JiangZJLRGT23,DBLP:conf/corl/LuoDWKGL23} that
state-conditioned decoding allows more compact codebooks while preserving reconstruction fidelity. For each token, our decoder conditions on $o_{t_b}$ and the quantized latent $\hat z^{(D)}_t$ via a linear adapter and causal attention, reconstructing all features for every token in the chunk by leveraging its previous context: $(\hat G_{t_{b-c}}, \hat G^{(L)}_{t_{b-c}}, \hat o_{t_{b-c}}, \hat m_{t_{b-c}}, \ldots, \hat G_{t_{b+1}}, \hat G^{(L)}_{t_{b+1}}, \hat o_{t_{b+1}}, \hat m_{t_{b+1}})$. We optimize a reconstruction loss plus a residual partial‑sum commitment:
\[
\mathcal{L}
= \sum_{\tau} \alpha_\tau \big\| (\hat x_\tau - x_\tau) \big\|_2^2
+ \frac{\beta_{\mathrm{ps}}}{D} \sum_{\ell=1}^{D} \big\| Z - \mathrm{sg}\!\left[\hat Z^{(\ell)}\right] \big\|_2^2.
\]
Here $\alpha_\tau=\alpha_{\text{tail}}$ for the last two tokens $(x_{t_b},x_{t_{b+1}})$ and $\alpha_{\text{ctx}}$ otherwise; $\hat Z^{(\ell)}$ is the depth-wise $\ell$ partial sum of residual code embeddings; and $\mathrm{sg}[\cdot]$ denotes stop-gradient. The depth‑wise partial‑sum term stabilizes residual quantization and prevents code hopping across depths, consistent with RQ-VAE~\citep{DBLP:conf/cvpr/LeeKKCH22}.

\subsection{Temporal Prior over Residual Code Stacks}
Let $k_{t,1:D}$ denote the depth-$1{:}D$ codes at macro time $t$. We learn an observation conditioned depth-aware autoregressive prior that factorizes across time and within-time depth:
\[
p_\phi\!\big(k_{t,1:D}\,\big|\,k_{<t,1:D},\,o_t\big)
\;=\;
\prod_{\ell=1}^{D}
p_\phi\!\big(k_{t,\ell}\,\big|\,k_{<t,1:D},\,k_{t,<\ell},\,o_t\big).
\]

Inspired by the factorized residual-code prior in RQ-Transformer~\citep{DBLP:conf/cvpr/LeeKKCH22}, we design an efficient prior for planning that avoids expanding a depth-$D$ code stack into $D$ separate tokens per time step (which would increase context length from $C$ to $CD$). Specifically, a temporal trunk (causal over $t$) embeds each past position by the sum of its depth embeddings $\sum_{j=1}^{D} e_{k_{u,j}}$, adds positional and observation embeddings, and produces a context $h_t$. We use a lightweight depth head (a shared Multi-Layer Perceptron) to predict $k_{t,1}, k_{t,2},\ldots,k_{t,D}$ by conditioning on $h_t$, a depth embedding, and the partial sum of shallower depths at time $t$. We then minimize the negative log-likelihood loss:
\[
\mathcal{L}_{\text{prior}}
\;=\;
\mathbb{E}\Big[-\sum_{t}\sum_{\ell=1}^{D} \log p_\phi\big(k_{t,\ell}\,\big|\,k_{<t,1:D},\,k_{t,<\ell},\,o_t\big)\Big].
\]
The time and depth factorization retains long temporal context while modeling within-position coarse to fine refinement efficiently.

\subsection{In-Context Planning With Monte Carlo Tree Search}
Prior work has leveraged MCTS to mitigate stochasticity arising from the environment in both online~\citep{DBLP:conf/iclr/AntonoglouSOHS22} and offline RL~\citep{luo2025scalable}. Planning in the real world, however, poses two additional challenges. First, partial observability induces apparent stochasticity when the context cannot reliably disambiguate latent states. Second, one needs a mechanism to balance the inherited bias and exploration at decision time when policies are learned from suboptimal behavior.


We therefore adopt MCTS as the online planner in our latent temporal-abstraction space (Fig.~\ref{fig:mcts}).
Using the learned latent dynamics to sample multiple future continuations and backing up their returns,
MCTS estimates the expected value of each candidate latent action, making action selection less sensitive
to noisy single-rollout return estimates. Moreover, by planning over a context-conditioned latent state,
the search mitigates apparent stochasticity from partial observability (to the extent captured by the context).
Finally, coupling P-UCT~\citep{silver2017masteringchessshogiselfplay} with our context-conditioned prior over latent tokens enables targeted exploration
and provides a principled way to override suboptimal priors when predicted returns justify it, while keeping the search
within the distribution supported by the learned model.

\begin{figure*}
    \centering
    \includegraphics[width=\columnwidth]{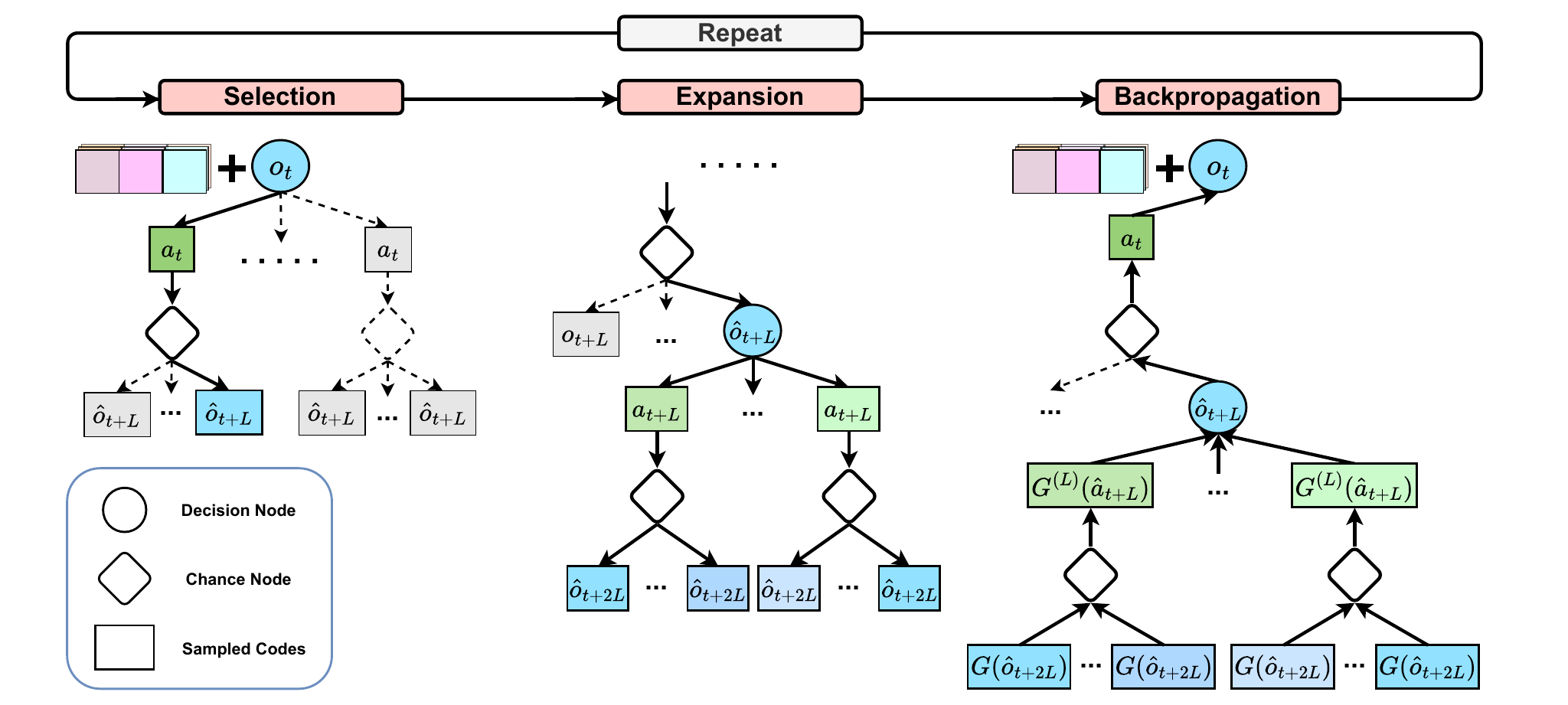}
    \caption{Macro-level MCTS overview. Each iteration uses P-UCT to select a macro-action, expands several candidates and their predicted outcomes in parallel, then backs up the resulting Q-estimates through the search tree to steer subsequent exploration.}
    \label{fig:model_training}
\end{figure*}\label{fig:mcts}

\paragraph{Latent Search Tree.}
At time $t$, the agent observes $o_t$ and an interaction history in a sliding window of length $L\!\times\!c$. We encode this history into residual-quantized codes to obtain a context window $k_{t-1:t-c,\,1:D}$. A \emph{decision node} is
$s=(o_t, k_{t-1:t-c,1:D})$, and an \emph{action edge} $a$ out of $s$ is a depth $D$ code stack $k_{t,1:D}$. Executing $a$ produces a
distribution over \emph{outcome codes} $k_{t+1,1:D}$ via our temporal prior
$p_\phi(k_{t+1,1:D} \mid k_{t,1:D}, o_t, k_{t-1:t-c,1:D})$, and each outcome is decoded
to a tail $(\hat{G}_{t+L},\hat{G}^{(L)}_{t+L}, \hat{o}_{t+L}, \hat{m}_{t+L})$, yielding the successor decision node $s_{t+L}$. To mitigate the cost of iterative model calls with context (a bottleneck when parallelism is underused), following~\citet{luo2025scalable} we pre-construct a
context-conditioned latent search tree by parallel sampling and caching a finite set of high-probability action stacks
and their predicted outcomes from $p_\phi$, and run MCTS over this cached search space (\cref{sec:algo}).


\paragraph{Policy-guided selection}
From node $s$, we restrict top-$K$ action candidates and use a behavior-like prior to prioritize searching
in-distribution actions without sacrificing exploration. Let the policy
head produce probability logits $l_a$,
we select $a$ by the AlphaZero-style P-UCT score:

\resizebox{\linewidth}{!}{$
a=\operatorname*{arg\,max}_{a}\Big[Q(s,a) \;+\;
\underbrace{\Big(c_1+\log\!\frac{N(s)+c_2+1}{c_2}\Big)\,
\frac{\sqrt{N(s)}}{1+N(s,a)}}_{\text{exploration term}}\,
\underbrace{\Big(\pi_T(a\mid s)=\frac{e^{l_a/T}}{\sum_{b\in\mathcal{A}(s)} e^{l_b/T}}\Big)}_{\text{temperature scaled prior}}\Big],
$}
where $N(s)$ and $N(s,a)$ are visit counts and ${Q}(s,a)$ is
the action value; $c_1,c_2>0$ are P-UCT exploration constants. This coupling of
a learned prior with P-UCT emphasizes in-distribution actions (large
$\pi_T$) without sacrificing exploration via the exploration term, allowing the search to deviate from the source policy when returns warrant it.

\paragraph{Parallel expansion and backpropagation.}
At each decision node $s_t$ we expand the top-$K$ candidates in parallel, sample $D$ outcome codes at the chance node, and decode them to obtain a successor $s_{t+L}\!=\!(\hat{o}_{t+L},\, k'_{t:t-c+1,\,1:D})$ and its leaf value. We then back up $Q(s,a)$ along the visited path using incremental averages and update visit counts.

\section{Experiments}
We evaluate {\itap} through comprehensive empirical studies using tasks from the D4RL benchmark~\citep{DBLP:journals/corr/abs-2004-07219}, focusing on standard Gym locomotion tasks and complex high-dimensional Adroit robotic manipulation. Our experiments assess both performance and adaptability of {\itap} across varying degrees of environmental stochasticity. We additionally conduct ablation studies to examine how macro-action length, context length, planning horizon, and residual depth affect performance. We further analyze the relationship between decision latency and context length in Section~\ref{sec:latency}.

\textbf{Baselines.} We compare {\itap} to strong offline RL baselines: model-free actor–critic methods Conservative Q-Learning (CQL; \citet{DBLP:conf/nips/KumarZTL20}) and Implicit Q-Learning (IQL; \citet{DBLP:conf/iclr/KostrikovNL22}); context-conditioned policy methods such as Decision Transformer (DT; \citet{DBLP:conf/nips/ChenLRLGLASM21}); and model-based planners that operate over learned temporal abstractions, including Trajectory Autoencoding Planner (TAP; \citet{DBLP:conf/iclr/JiangZJLRGT23}), Latent Macro-Action Planner (L-MAP; \citet{luo2025scalable}), where TAP is largely insensitive to raw action dimensionality and shows strong performance on high-dimensional Adroit manipulation, and L‑MAP likewise scales well to high‑dimensional control while remaining robust under stochastic dynamics. Finally, a risk-sensitive, model-based specialist for stochastic domains (1R2R; \citet{DBLP:conf/nips/RigterLH23}). 

\textbf{Experimental Setup.} To assess {\itap}'s adaptation capabilities across varying latent task parameters and associated dynamics, we conduct comprehensive experiments using the Stochastic MuJoCo tasks introduced by~\citet{DBLP:conf/nips/RigterLH23}. Each environment defines an unobserved global latent task parameter controlling perturbation levels (0, 2.5, 5), which in turn specify the distribution of instantaneous hidden forces at each step. We train a single model per method (\itap, DT, L‑MAP) with three random seeds on the pooled data across perturbation regimes for each behavior-policy dataset type. At test time, \itap{} and DT perform no gradient updates; adaptation arises only through conditioning on the context window. We also train L-MAP on the same pooled data as a planning baseline that is robust to regime variation but does not rely on in-context adaptation. For standard offline RL baselines (e.g., CQL/IQL), we compare the methods with dataset-specific models. To test scalability to high‑dimensional control and partial observability, we evaluate on Adroit~\citep{DBLP:conf/rss/RajeswaranKGVST18} in (i) the original fully observable setting and (ii) a partially observable variant where we mask a subset of target‑position coordinates (see Appendix~\ref{sec:experiment_domains} for the exact dimensions and masking procedure). Unless otherwise noted, we set the latent context size to $C=6$ tokens; with a macro length of $L=3$, this window summarizes 18 past primitive transitions. Further domain and hyperparameter details appear in Appendix~\ref{sec:experiment_domains}.

\subsection{Main results}\label{sec:results}

\begin{table}[htbp]
  \centering
  \caption{Normalised results for high-noise (5), moderate-noise (2.5), deterministic (0) environments. Bold numbers indicate the best scores in each row.}
  \label{tab:combined_noise}
  \resizebox{\linewidth}{!}{%
    \begin{tabular}{llccccccc}
      \toprule
      \multicolumn{2}{c}{} &
        \multicolumn{4}{c}{\textbf{Model-Based}} &
        \multicolumn{3}{c}{\textbf{Model-Free}}\\
      \cmidrule(lr){3-6}\cmidrule(lr){7-9}
      \textbf{Dataset Type} & \textbf{Env (Noise)} &
        \textbf{\itap} & \textbf{L-MAP} & \textbf{TAP} & \textbf{1R2R} &
        \textbf{DT} & \textbf{CQL} & \textbf{IQL}\\
      \midrule
      Medium-Expert & Hopper (5) & \textbf{82.87 $\pm$ 3.55} & 71.49 $\pm$ 3.46 & 37.31 $\pm$ 3.66 & 37.99 $\pm$ 2.71 & 61.87 $\pm$ 2.56 & 68.03 $\pm$ 3.94 & 44.83 $\pm$ 2.58\\
                    & Hopper (2.5)  & 104.45 $\pm$ 2.66 & 93.40 $\pm$ 3.65 & 40.86 $\pm$ 5.42 & 52.19 $\pm$ 8.37 & 73.86 $\pm$ 2.68 & \textbf{106.17 $\pm$ 2.16} & 60.61 $\pm$ 3.46\\
                    & Hopper (0)  & \textbf{111.71 $\pm$ 0.08} & 106.74 $\pm$ 2.24 & 85.55 $\pm$ 3.83 & 57.40 $\pm$ 6.06 & 101.6 $\pm$ 1.85 & 105.4 & 91.5\\
                    & Walker2D (12)& \textbf{93.50 $\pm$ 3.15} & 92.75 $\pm$ 1.34 & 91.09 $\pm$ 2.78 & 32.38 $\pm$ 4.55 & 52.42 $\pm$ 1.27 & 83.18 $\pm$ 3.70 & 68.61 $\pm$ 3.33\\
                    & Walker2D (7)& \textbf{97.29 $\pm$ 1.68} & 93.48 $\pm$ 1.20 & 91.40 $\pm$ 1.42 & 56.48 $\pm$ 7.51 & 64.67 $\pm$ 1.00 & 91.44 $\pm$ 1.44 & 86.66 $\pm$ 1.84\\
                    & Walker2D (0)& 101.20 $\pm$ 1.91 & 100.38 $\pm$ 0.72 & 105.32 $\pm$ 2.03 & 73.18 $\pm$ 6.29 & 64.65 $\pm$ 0.79 & 108.8 & \textbf{109.6}\\
      \cmidrule(lr){1-9}
      \multicolumn{2}{l}{Mean (Medium-Expert)} &
        \textbf{98.50} & 93.04 & 75.26 & 51.60 & 69.85 & 93.84 & 76.97\\
      \midrule
      Medium        & Hopper (5) & \textbf{67.80 $\pm$ 2.60} & 59.05 $\pm$ 2.93 & 43.93 $\pm$ 2.66 & 33.99 $\pm$ 0.92 & 55.91 $\pm$ 2.02 & 45.21 $\pm$ 2.97 & 49.69 $\pm$ 2.47\\
                    & Hopper (2.5)  & \textbf{72.47 $\pm$ 2.71} & 63.21 $\pm$ 3.10 & 43.64 $\pm$ 2.25 & 65.24 $\pm$ 3.31 & 60.97 $\pm$ 0.82 & 49.92 $\pm$ 3.00 & 56.00 $\pm$ 3.60\\
                    & Hopper (0)  & \textbf{81.94 $\pm$ 2.14} & 61.65 $\pm$ 2.81 & 69.14 $\pm$ 2.33 & 55.49 $\pm$ 3.99 & 58.14 $\pm$ 0.24 & 58.0 & 66.3\\
                    & Walker2D (12)& 60.35 $\pm$ 2.73 & 59.05 $\pm$ 2.30 & 52.20 $\pm$ 2.76 & 32.13 $\pm$ 4.51 & 32.20 $\pm$ 0.83 & \textbf{61.49 $\pm$ 3.24} & 47.53 $\pm$ 3.05\\
                    & Walker2D (7)& \textbf{65.85 $\pm$ 2.44} & 62.23 $\pm$ 1.84 & 44.46 $\pm$ 1.82 & 65.16 $\pm$ 2.84 & 43.77 $\pm$ 0.95 & 49.38 $\pm$ 2.02 & 48.82 $\pm$ 2.31\\
                    & Walker2D (0)& \textbf{79.57 $\pm$ 1.22} & 75.54 $\pm$ 1.59 & 51.75 $\pm$ 3.30 & 55.69 $\pm$ 4.97 & 55.36 $\pm$ 0.61 & 72.5 & 78.3\\
      \cmidrule(lr){1-9}
      \multicolumn{2}{l}{Mean (Medium)} &
        \textbf{71.33} & 63.46 & 50.85 & 51.28 & 51.06 & 56.08 & 57.77\\
      \midrule
      Medium-Replay & Hopper (5) &  \textbf{70.67 $\pm$ 2.59} & 60.76 $\pm$ 2.79 & 48.69 $\pm$ 2.97 & 68.25 $\pm$ 3.78 & 35.17 $\pm$ 0.96 & 51.70 $\pm$ 3.09 & 43.27 $\pm$ 2.78\\
                    & Hopper (2.5)  &   \textbf{81.33 $\pm$ 2.19} & 73.81 $\pm$ 2.67 & 38.10 $\pm$ 3.22 & 22.82 $\pm$ 2.08 & 35.76 $\pm$ 1.01 & 40.53 $\pm$ 1.52 & 49.12 $\pm$ 3.38\\
                    & Hopper (0)  & 86.57 $\pm$ 2.03 & 90.8 $\pm$ 0.63 & 80.92 $\pm$ 3.79 & 89.67 $\pm$ 1.92 & 43.01 $\pm$ 1.36 & \textbf{95.0} & 94.7\\
                    & Walker2D (12)& \textbf{70.60 $\pm$ 3.07} & 59.16 $\pm$ 2.92 & 55.15 $\pm$ 3.29 & 65.63 $\pm$ 3.41 & 37.22 $\pm$ 0.78 & 50.33 $\pm$ 3.88 & 45.13 $\pm$ 2.38\\
                    & Walker2D (7)& \textbf{74.04 $\pm$ 1.94} & 69.20 $\pm$ 2.55 & 43.49 $\pm$ 2.27 & 52.23 $\pm$ 2.22 & 49.51 $\pm$ 0.81 & 40.24 $\pm$ 1.67 & 40.77 $\pm$ 2.72\\
                    & Walker2D (0)&  75.49 $\pm$ 2.52 & 70.66 $\pm$ 1.78 & 72.32 $\pm$ 3.26 & \textbf{90.67 $\pm$ 1.98} & 48.44 $\pm$ 0.76 & 77.2 & 77.2\\
      \cmidrule(lr){1-9}
      \multicolumn{2}{l}{Mean (Medium-Replay)} &
        \textbf{76.45} & 70.73 & 56.45 & 64.88 & 41.52 & 59.17 & 58.37\\
      \bottomrule
    \end{tabular}%
  }
\end{table}

\textbf{Mujoco} We use the Stochastic MuJoCo suite to evaluate \itap{}'s in-context adaptation and robustness to uncertainty. Each episode is governed by a latent task parameter that selects the perturbation regime (e.g., 0, 2.5, 5) and thereby determines the distribution of hidden forces. We therefore report results in two complementary settings. 
First, Table~\ref{tab:combined_noise} evaluates \emph{in-distribution} regimes where latent-parameter values that are present in the meta-training datasets so that performance primarily reflects a method's ability to infer the active regime from history, adapt its decisions online, and handle uncertainty. Second, Table~\ref{tab:extrapolation} evaluates held-out regimes not included in training to evaluate interpolation and extrapolation across latent task parameters.

Across all dataset types, \itap{} achieves the highest mean score and remains strong across changing dynamics without any gradient updates at test time. These gains are consistent with \itap{}'s ability to (i) leverage trajectory history for in-context identification of the latent regime and (ii) plan online to optimize actions instead of merely imitating the behavior policy in the dataset. Relative to L-MAP, \itap{}'s improvements highlight the benefit of \emph{context-guided} action selection within search: the sequence-model prior steers exploration while the planner evaluates alternatives, rather than relying on MCTS alone to handle uncertainty. In contrast, DT degrades more sharply under stochastic dynamics and lower-quality behavior data, consistent with its lack of an integrated planner and its reliance on return conditioning, which can be sensitive to luck-induced variability~\citep{DBLP:conf/nips/PasterMB22}. By combining context-based adaptation with downstream planning, \itap{} is less likely to inherit suboptimal dataset behavior and can deviate whenever the planner identifies more promising actions.

\begin{table}[t]
\centering
\caption{Normalised results (mean $\pm$ std). Noise parameter values are shown in parentheses. Bold indicates the best score in each row.}
\label{tab:extrapolation}
\begin{tabular}{llccc}
\toprule
Dataset Type & Env (Noise) & I-TAP & L-MAP & DT \\
\midrule

\multirow{6}{*}{Medium-Expert}
& Hopper (7.5)   & \textbf{70.43 $\pm$ 3.17}  & 62.90 $\pm$ 3.50  & 58.96 $\pm$ 0.21 \\
& Hopper (3.75)  & \textbf{85.07 $\pm$ 3.57}  & 80.51 $\pm$ 3.86  & 64.97 $\pm$ 0.25 \\
& Hopper (1.25)  & \textbf{106.91 $\pm$ 2.17} & 94.65 $\pm$ 3.03  & 87.41 $\pm$ 0.24 \\
& Walker2D (15)  & \textbf{88.26 $\pm$ 3.67}  & 87.21 $\pm$ 3.42  & 44.07 $\pm$ 0.26 \\
& Walker2D (9.5) & \textbf{94.29 $\pm$ 1.97}  & 92.31 $\pm$ 1.81  & 56.23 $\pm$ 0.22 \\
& Walker2D (3.5) & \textbf{98.80 $\pm$ 1.00}  & 96.53 $\pm$ 0.62  & 65.07 $\pm$ 0.18 \\
\midrule
\multicolumn{2}{l}{Mean (Medium-Expert)} & \textbf{90.63} & 85.69 & 62.79 \\
\midrule

\multirow{6}{*}{Medium}
& Hopper (7.5)   & \textbf{60.88 $\pm$ 2.49} & 55.26 $\pm$ 2.53 & 51.62 $\pm$ 0.16 \\
& Hopper (3.75)  & \textbf{71.43 $\pm$ 2.75} & 66.68 $\pm$ 2.50 & 56.09 $\pm$ 0.17 \\
& Hopper (1.25)  & \textbf{74.60 $\pm$ 2.46} & 71.08 $\pm$ 2.56 & 59.39 $\pm$ 0.12 \\
& Walker2D (15)  & \textbf{54.60 $\pm$ 2.52} & 50.10 $\pm$ 2.82 & 30.69 $\pm$ 0.18 \\
& Walker2D (9.5) & \textbf{60.22 $\pm$ 2.49} & 55.96 $\pm$ 2.53 & 39.16 $\pm$ 0.19 \\
& Walker2D (3.5) & \textbf{76.56 $\pm$ 1.58} & 75.77 $\pm$ 1.59 & 51.89 $\pm$ 0.17 \\
\midrule
\multicolumn{2}{l}{Mean (Medium)} & \textbf{66.38} & 62.47 & 48.14 \\
\midrule

\multirow{6}{*}{Medium-Replay}
& Hopper (7.5)   & \textbf{56.96 $\pm$ 3.07} & 50.79 $\pm$ 2.56 & 30.17 $\pm$ 0.14 \\
& Hopper (3.75)  & \textbf{75.22 $\pm$ 2.72} & 70.06 $\pm$ 2.87 & 33.03 $\pm$ 0.18 \\
& Hopper (1.25)  & \textbf{84.43 $\pm$ 2.09} & 83.99 $\pm$ 2.21 & 39.22 $\pm$ 0.24 \\
& Walker2D (15)  & \textbf{60.70 $\pm$ 3.11} & 50.64 $\pm$ 2.87 & 29.91 $\pm$ 0.15 \\
& Walker2D (9.5) & \textbf{71.90 $\pm$ 2.22} & 64.68 $\pm$ 2.71 & 42.78 $\pm$ 0.16 \\
& Walker2D (3.5) & \textbf{75.11 $\pm$ 1.86} & 71.98 $\pm$ 2.12 & 58.49 $\pm$ 0.15 \\
\midrule
\multicolumn{2}{l}{Mean (Medium-Replay)} & \textbf{70.72} & 65.36 & 38.93 \\
\bottomrule
\end{tabular}
\end{table}

\begin{table}[htbp]
  \centering
  \caption{Adroit robotic hand control results.}
  \label{tab:results3}
  \begin{tabular}{llccc}
    \toprule
    \textbf{Dataset Type} & \textbf{Env} & \textbf{\itap} & \textbf{L-MAP} & \textbf{TAP}\\
    \midrule
    Cloned  & Pen      & \textbf{85.44 $\pm$ 8.19} & 60.68 $\pm$ 7.88 & 46.44 $\pm$ 7.54\\
    Cloned  & Hammer   & \textbf{4.38 $\pm$ 1.28}  & 2.43  $\pm$ 0.29 & 1.32  $\pm$ 0.12\\
    Cloned  & Door     & \textbf{14.17 $\pm$ 1.34} & 13.22 $\pm$ 1.34 & 13.45 $\pm$ 1.43\\
    Cloned  & Relocate & 0.08  $\pm$ 0.02          & \textbf{0.15  $\pm$ 0.13}& -0.23 $\pm$ 0.01\\
    \midrule
    Expert  & Pen      & \textbf{133.81 $\pm$ 5.23} & 126.60 $\pm$ 5.60  & 127.40 $\pm$ 7.70\\
    Expert  & Hammer   & \textbf{128.37 $\pm$ 0.21} & 127.16 $\pm$ 0.29  & 127.60 $\pm$ 1.70\\
    Expert  & Door     & \textbf{105.98 $\pm$ 0.08} & 105.24 $\pm$ 0.10  & 104.80 $\pm$ 0.80\\
    Expert  & Relocate & \textbf{109.85 $\pm$ 0.88} & 107.57 $\pm$ 0.76  & 106.21 $\pm$ 1.61 \\
    \midrule
    Expert (POMDP)  & Pen      & \textbf{81.68 $\pm$ 9.83}  & 69.84 $\pm$ 9.81  & 60.87 $\pm$ 9.55\\
    Expert (POMDP)  & Hammer   & \textbf{72.36 $\pm$ 8.48}  & 59.21 $\pm$ 6.52  & 42.22 $\pm$ 12.92\\
    Expert (POMDP)  & Door     & \textbf{95.05 $\pm$ 3.12}  & 89.35 $\pm$ 3.41  & 83.71 $\pm$ 4.22\\
    Expert (POMDP)  & Relocate & \textbf{52.16 $\pm$ 4.46}  & 37.36 $\pm$ 3.84  & 33.94 $\pm$ 3.50\\
    \midrule
    \textbf{Mean (Expert)} & & \textbf{119.50} & 116.64 & 116.50\\
    \textbf{Mean (Expert POMDP)} & & \textbf{75.31} & 63.94 & 55.19\\
    \bottomrule
  \end{tabular}
\end{table}

\textbf{Adroit Control}  
Adroit poses high‑dimensional state–action spaces and fine‑grained control demands. Table~\ref{tab:results3} shows that \itap{} achieves the best performance in both fully observable and partially observable setting. On cloned datasets, \itap{} outperforms both L‑MAP and TAP in three of four tasks, indicating that planning with context-guided prior in the latent abstraction space enables the agent to do more targeted exploration with promising returns while remaining in distribution. On expert datasets, \itap{} matches or exceeds baselines, and under partial observability (POMDP) it maintains the top scores across all four tasks, highlighting the benefit of context‑conditioned planning when state aliasing induces apparent stochasticity and its ability to leverage feedback from the environment to provide more targeted exploration to improve decisions. We use residual depth D=2 for cloned and D=3 for expert datasets; the stronger expert results are consistent with the intuition that coarse‑to‑fine residual quantization helps retain the granularity needed for high-dimensional continuous control.

\subsection{Ablation Study}\label{sec:ablation}
\begin{figure*}
    \centering
    \includegraphics[width=\columnwidth]{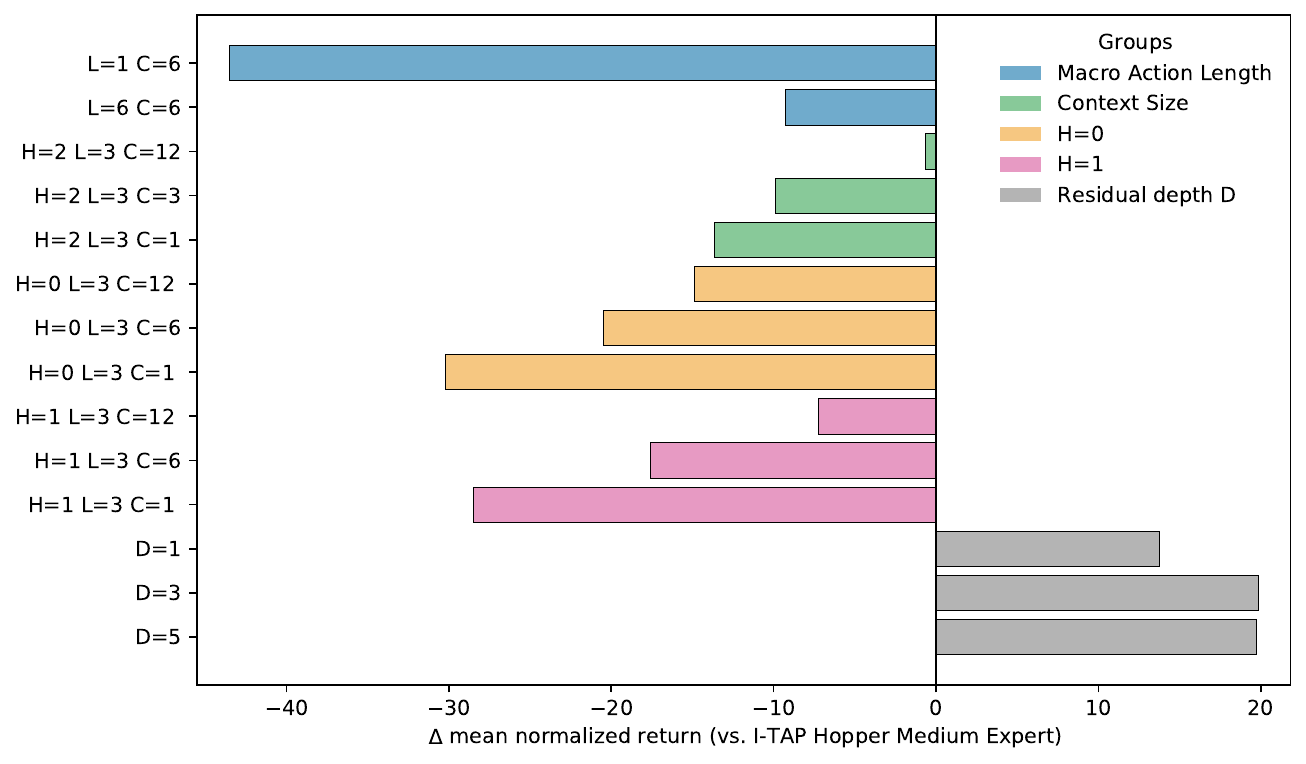}
    \caption{Ablation results across Adroit (expert) and MuJoCo Hopper. We plot $\Delta$ scores relative to \itap{} on Hopper medium\textendash expert, which serves as the baseline (zero line).}
    \label{fig:ablation}
\end{figure*}

We present analyses and ablations of macro action length, context length, planning horizon, and residual depth.
Figure~\ref{fig:ablation} summarizes the results from ablation studies conducted across deterministic, moderate-noise, and high-noise Mujoco Hopper control tasks and Adroit (expert) tasks.

\textbf{Macro Action Length.} 
We vary the macro length \(L\) and compare \(L{=}1\) versus \(L{=}6\). Increasing to \(L{=}6\) does \emph{not} noticeably degrade performance, whereas decreasing to \(L{=}1\) causes a substantial drop. Two factors explain this: (i) with fixed \(C\), the \emph{effective history} observed by the planner scales as \(C \!\times\! L\) steps, so \(L{=}1\) shortens the usable history and weakens in\mbox{-}context adaptation; and (ii) shorter macros increase the \emph{branching factor} during search and make the RQ\mbox{-}VAE more prone to \emph{overfitting}, which requires careful hyperparameter control. 

\textbf{Context Size.} 
We ablate context size \(C \in \{1,3,12\}\) (in latent tokens). The datasets are produced under distinct latent dynamics; thus, context is useful for \emph{inferring the active mode} and for feeding back recent rewards to steer exploration. Increasing context from \(C{=}1\) (approximately \(3\) past transitions when \(L{=}3\)) to \(C{=}12\) (approximately \(36\) transitions) yields consistent gains. This demonstrates that short, latent contexts already help (few tokens cover many steps), and performance improves as \(C\) grows because the model better disambiguates latent parameters and aggregates noisy feedback. 



\textbf{Planning Horizon.} We vary the MCTS look\mbox{-}ahead depth \(H\) (in latent tokens). With macro length \(L{=}3\), a depth \(H\) corresponds to \(H\!\times\!L\) planning horizon in the raw action space. Removing planning (\(H{=}0\)) harms performance for all \(C\); the drop is smaller on deterministic tasks where the in\mbox{-}context prior is already strong, but it remains insufficient under stochasticity. Increasing to \(H{=}1\) yields consistent gains across settings; at \(H{=}2\), we observe a pronounced jump, a shorter\mbox{-}context model (\(C{=}6\)) matches a longer\mbox{-}context baseline (\(C{=}12\)). This supports our hypothesis that MCTS could \emph{mitigates uncertainty} due to partial observability and stochastic dynamics by taking expectations over futures: deeper look\mbox{-}ahead can substitute for additional context up to a point before compute and diminishing returns dominate.



\begin{figure*}[t]
  \centering
  \begin{subfigure}{.24\textwidth}
    \includegraphics[width=\linewidth]{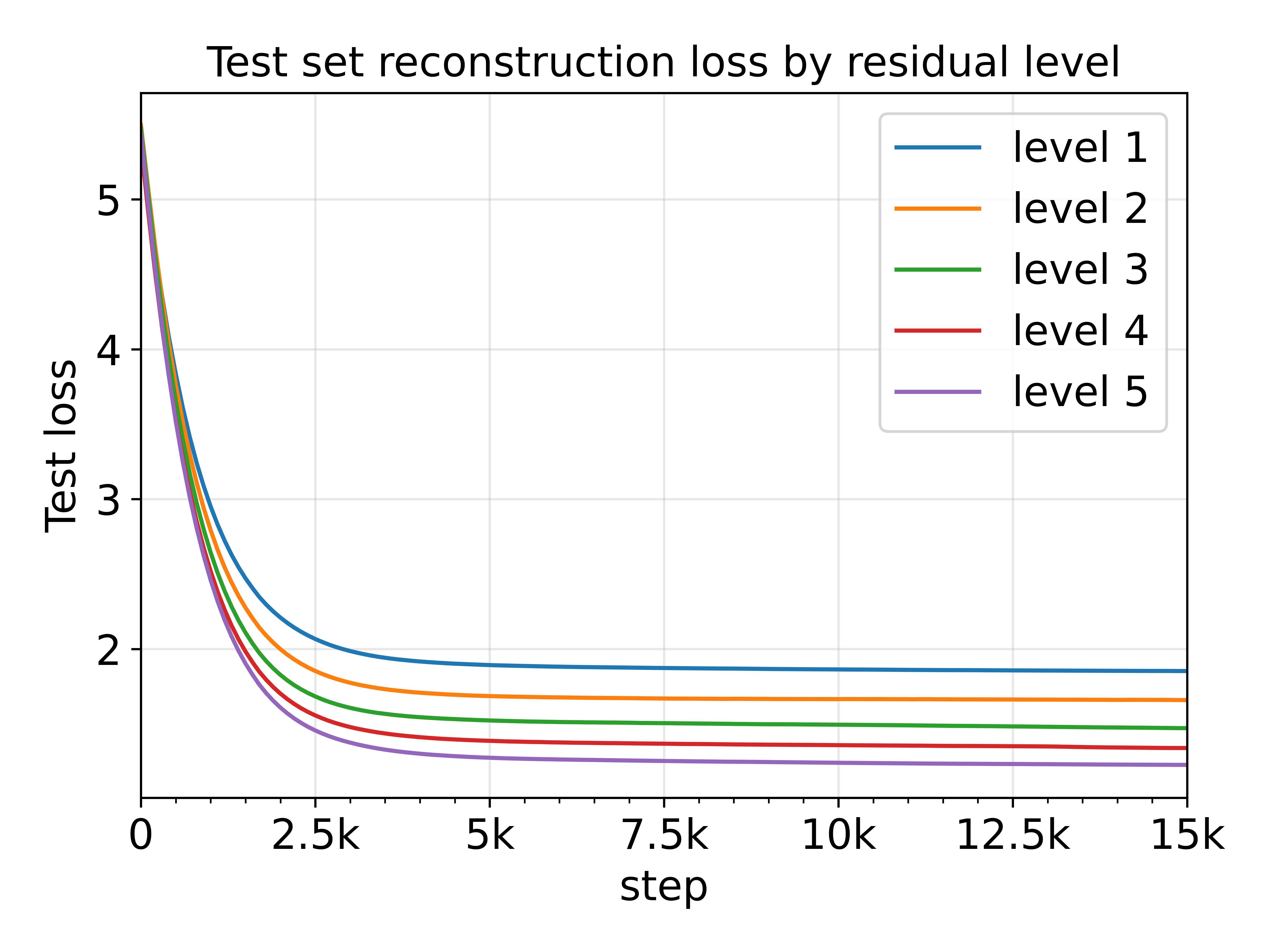}
    \caption{Hammer}
  \end{subfigure}
  \begin{subfigure}{.24\textwidth}
    \includegraphics[width=\linewidth]{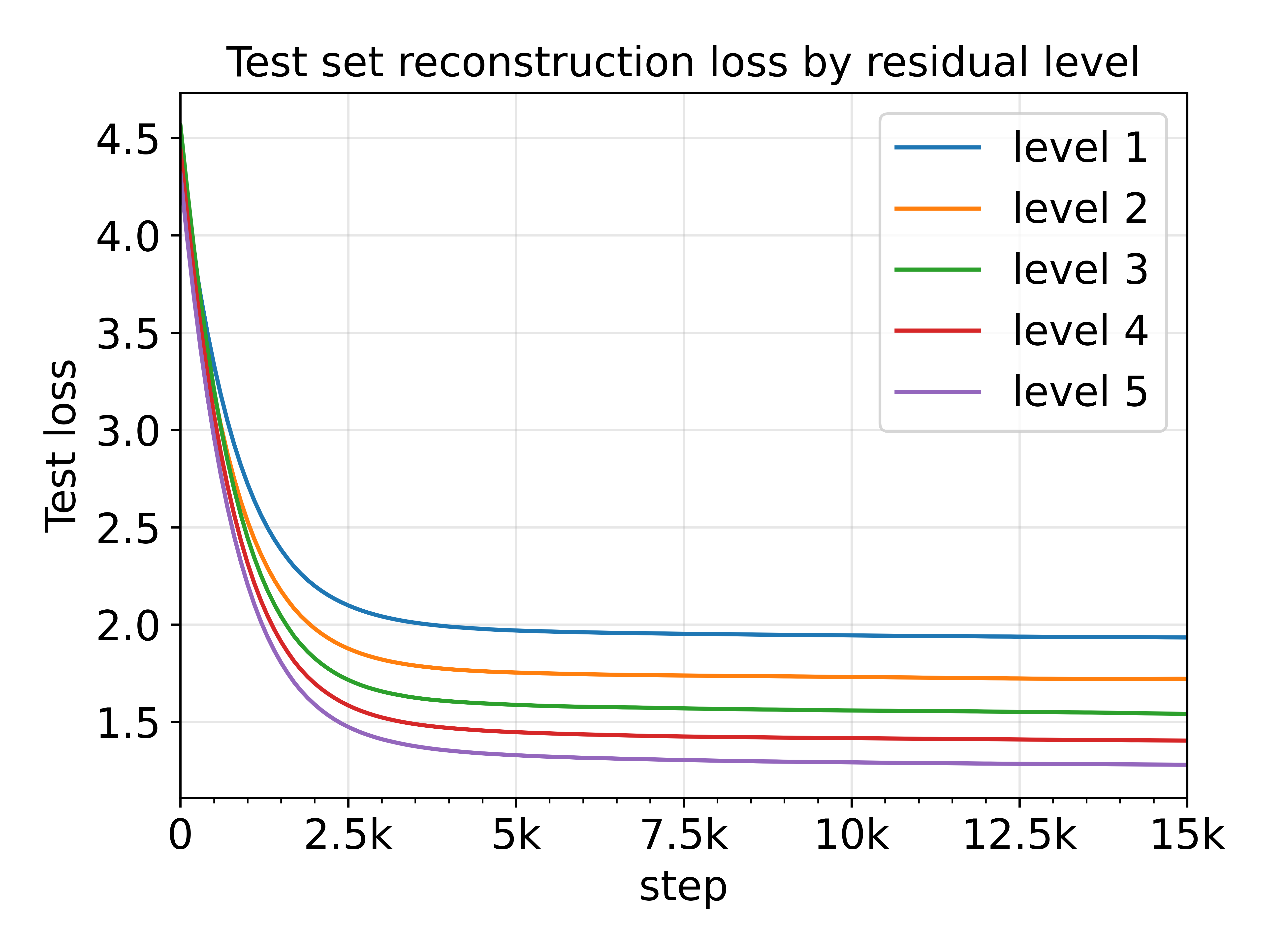}
    \caption{Door}
  \end{subfigure}
  \begin{subfigure}{.24\textwidth}
    \includegraphics[width=\linewidth]{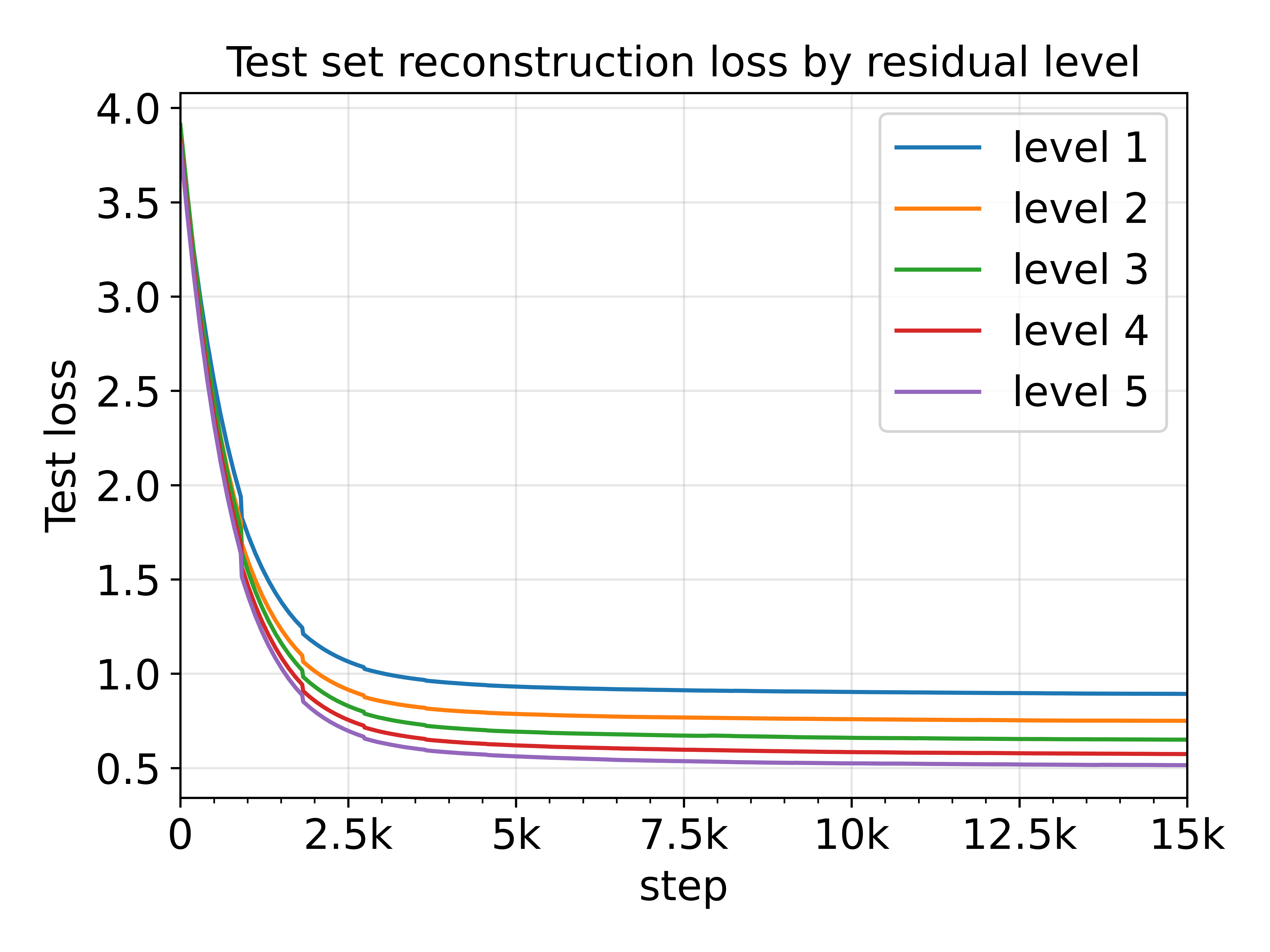}
    \caption{Pen}
  \end{subfigure}
  \begin{subfigure}{.24\textwidth}
    \includegraphics[width=\linewidth]{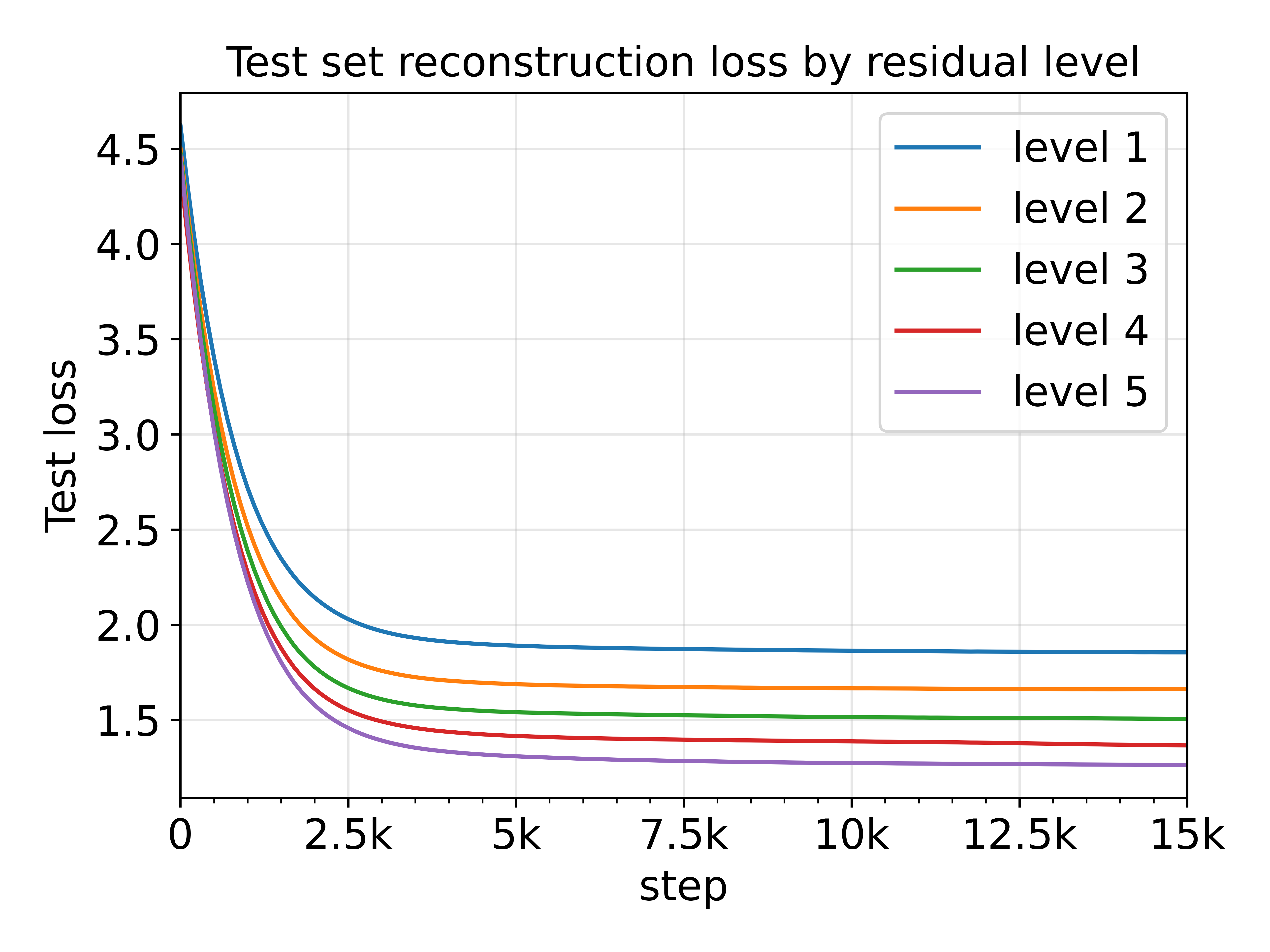}
    \caption{Relocate}
  \end{subfigure}
  \caption{Test reconstruction losses across residual levels.}\label{fig:reconstruction}
\end{figure*}

\textbf{Residual Depth}
We vary the residual quantization depth \(D\) for RQ\mbox{-}VAE. Reconstruction error \emph{decreases monotonically} with \(D\) (Figure~\ref{fig:reconstruction}), with diminishing returns: the largest drop is from \(D{=}1\) to \(D{=}3\). Control performance mirrors this: the task\mbox{-}averaged score rises from 113.40 at \(D{=}1\) to 119.50 at \(D{=}3\), and then plateaus (119.37 at \(D{=}5\)). This highlights that increasing residual depth improves reconstruction quality for high-dimensional features and shows that \(D{=}3\) preserves action granularity for high\mbox{-}DoF control without incurring the computation/optimization overhead of very deep stacks.

\subsection{Decision Time vs Context Length}\label{sec:latency}
\begin{figure}[htbp]
\centering
\includegraphics[width=0.45\textwidth]{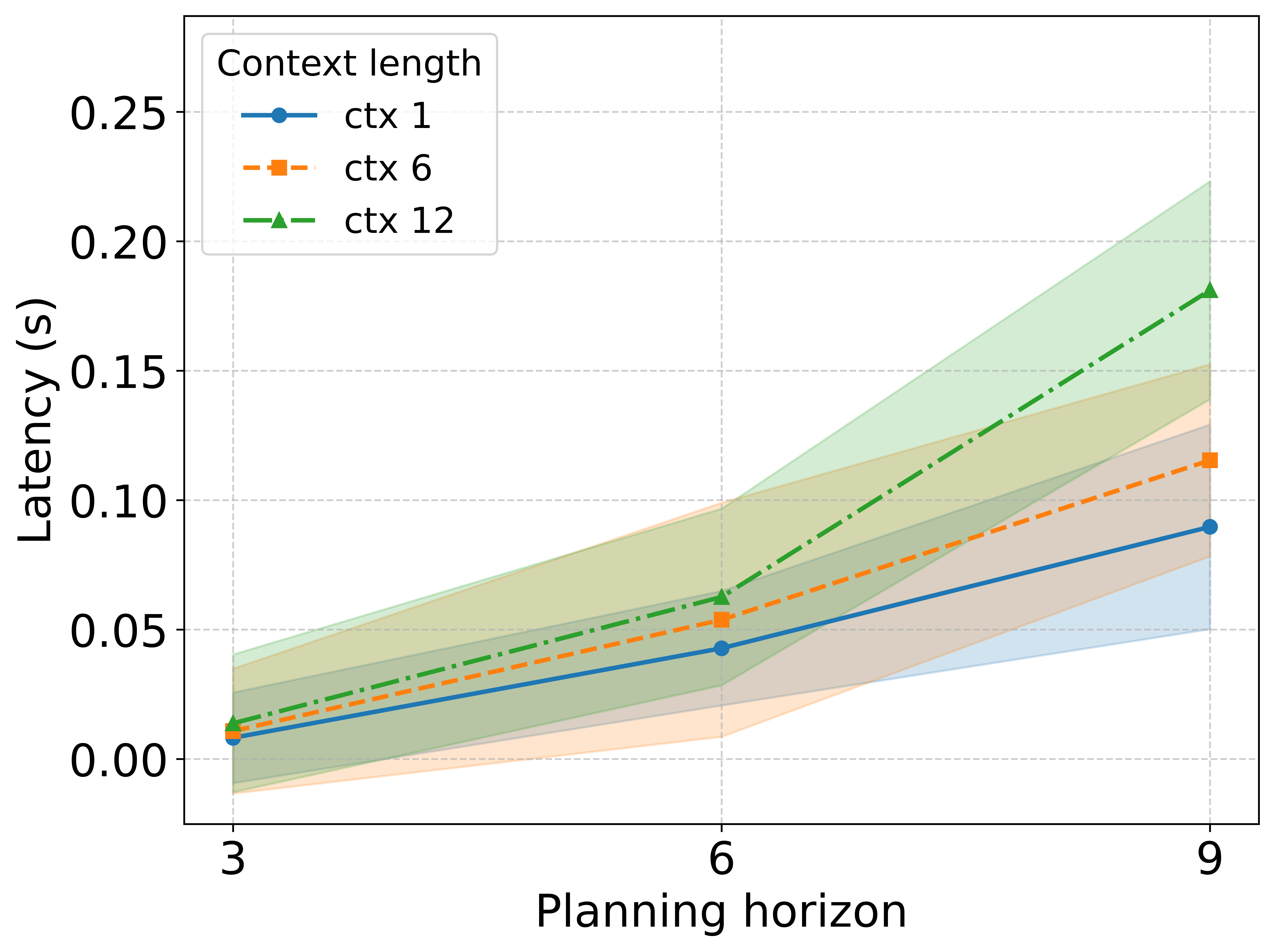}
\includegraphics[width=0.45\textwidth]{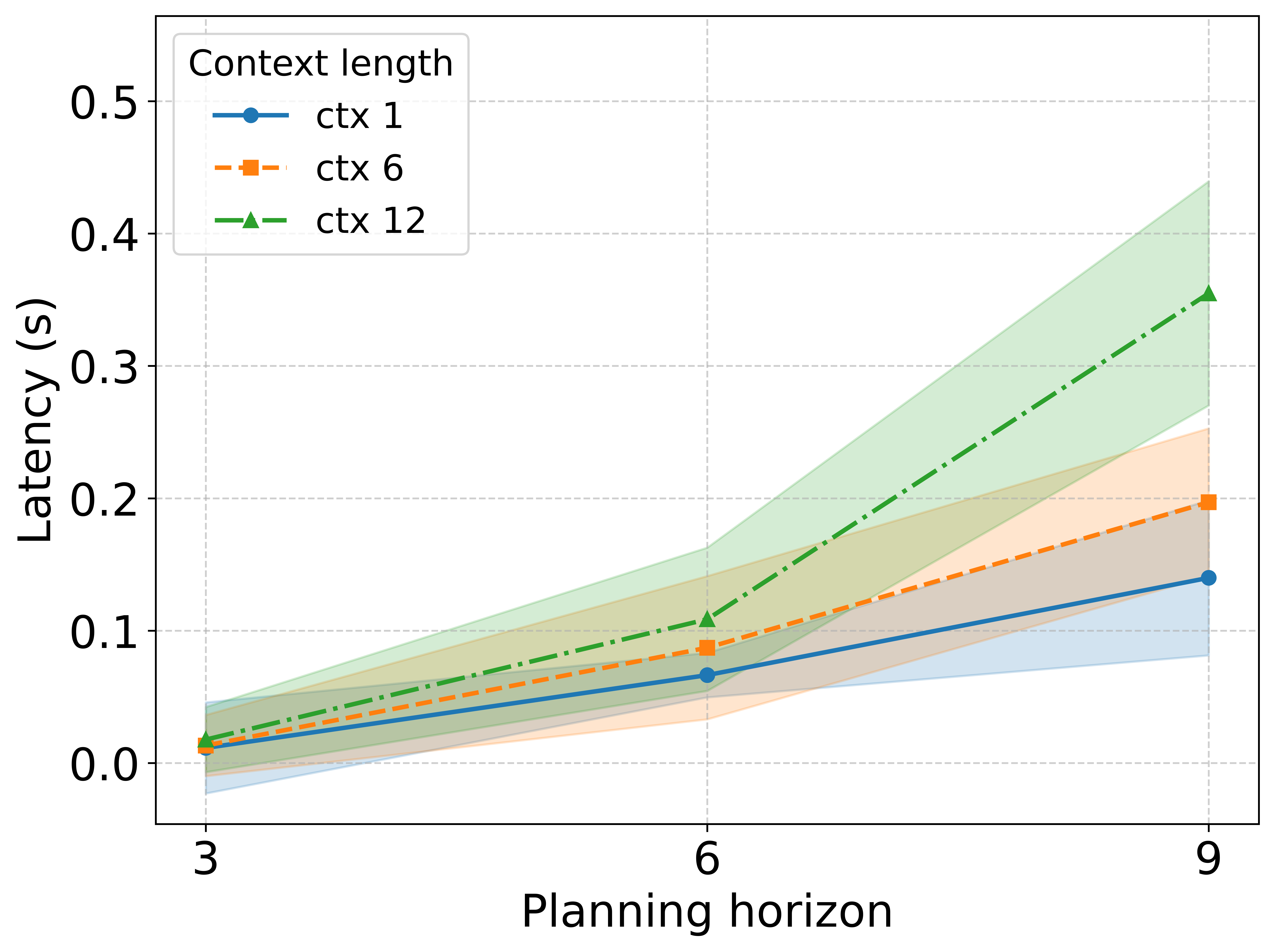}
\caption{Average decision time vs. context length with 16 action candidates (left) and 32 action candidates (right).}
\label{fig:avg_decision_time_bs16}
\label{fig:avg_decision_time_bs32}
\end{figure}

The latency results clearly highlight the trade-off between context size, planning depth, and decision-making efficiency. Our analysis of decision latency versus context length reveals that longer contexts substantially increase the average decision time, particularly as the planning depth increases. For instance, at planning horizon 9, decision latency escalates sharply from approximately 0.09 seconds (context size 1) to 0.35 seconds (context size 12) when the number of action candidates is 32. Similarly, at lower planning depths (1 and 2), longer contexts still cause noticeable latency increases, albeit less dramatically.

Thus, in scenarios characterized by dynamics influenced by unobserved states, our findings underline the necessity of carefully selecting context lengths. It is crucial to encode historical observations effectively, extracting the most informative signals without redundant details that contribute minimal additional predictive power but significantly raise computational latency. In practical applications, particularly those involving stochasticity and continuous observation-action spaces, balancing sufficient context length to maintain accuracy against the latency imposed by deeper planning becomes a critical design decision. Efficient decision-making, therefore, relies on identifying a context size that captures the essential dynamics without incurring unnecessary computational overhead.

\section{Related Works}

\paragraph{Reinforcement Learning as Sequence Modeling.} Recent advances have reframed reinforcement learning (RL) as a sequence modeling problem, initiated by Decision Transformer (DT), which formulates RL as supervised learning conditioned on desired returns~\citep{DBLP:conf/nips/ChenLRLGLASM21}. Building on these ideas, Algorithm Distillation~\citep{DBLP:conf/iclr/LaskinWOPSSSHFB23} and Decision-Pretrained Transformer~\citep{DBLP:conf/nips/0002XPCFNB23} leverage Transformers to distill optimal behaviors from historical trajectories, enabling rapid in-context adaptation. More recently, hierarchical variants like In-context Decision Transformer (IDT; \citet{DBLP:conf/icml/HuangHCS024}) extend this paradigm by modeling high-level decisions, alleviating computational bottlenecks associated with long context windows. However, as noted by~\citet{son2025distilling}, these methods may replicate suboptimal behaviors due to the absence of explicit planning mechanisms, a limitation potentially addressed by model-based planning. Moreover, supervised RL methods typically assume deterministic or near-deterministic datasets, inherently limiting their effectiveness in stochastic environments, where conditioning solely on outcomes can lead to incorrect decisions~\citep{DBLP:conf/nips/PasterMB22}. Unlike prior works relying on large contexts in near-deterministic settings, our approach explicitly targets efficient adaptation in dynamic, stochastic environments using a limited context window.

\paragraph{Model-Based Reinforcement Learning.}
From a model-based perspective~\citep{DBLP:conf/iclr/AntonoglouSOHS22, DBLP:journals/nature/SchrittwieserAH20}, \citet{DBLP:conf/nips/JannerLL21} introduced beam search over a Transformer dynamics model for planning, inspiring subsequent methods that plan in learned latent spaces. In particular, TAP~\citep{DBLP:conf/iclr/JiangZJLRGT23} and L-MAP~\citep{luo2025scalable} employ temporal abstraction by encoding multi-step action segments into discrete codes via state-conditioned VQ-VAEs, then planning over these compact tokens. While beam search with a learned model is effective in largely deterministic settings, L-MAP further adopts MCTS to handle stochastic dynamics and improve robustness. Nevertheless, both methods assume full observability, which can limit performance under state aliasing. Our proposed \itap{} bridges this gap by conditioning planning on recent histories to mitigate partial observability and use MCTS to take expectations over possible futures and deviate from suboptimal priors. 

\section{Discussion}
We presented the \textit{In-Context Latent Temporal Abstraction Planner} (\itap{}), an offline RL approach that discretizes observation--macro-action trajectories and performs planning in a learned latent space.
By combining temporal abstraction with in-context conditioning, \itap{} reduces planning complexity while enabling adaptation to latent regime variations through the test-time context window.
Residual quantization provides a coarse-to-fine code stack to extend the scalability of the proposed method, and our factorized prior keeps planning-time inference efficient when the model is queried.
Across stochastic MuJoCo and high-dimensional Adroit domains, \itap{} achieves competitive performance relative to strong offline baselines.
More broadly, the same idea of discretized latent representations plus planning-compatible sequence priors may be useful beyond offline RL, for example in imitation learning settings that require temporally extended actions while retaining fine-grained control.


\bibliography{references}

@article{SUTTON1999181,
title = {Between MDPs and semi-MDPs: A framework for temporal abstraction in reinforcement learning},
journal = {Artificial Intelligence},
volume = {112},
number = {1},
pages = {181-211},
year = {1999},
issn = {0004-3702},
doi = {https://doi.org/10.1016/S0004-3702(99)00052-1},
url = {https://www.sciencedirect.com/science/article/pii/S0004370299000521},
author = {Richard S. Sutton and Doina Precup and Satinder Singh}
}

@article{DBLP:journals/corr/abs-2004-07219,
  author       = {Justin Fu and
                  Aviral Kumar and
                  Ofir Nachum and
                  George Tucker and
                  Sergey Levine},
  title        = {{D4RL:} Datasets for Deep Data-Driven Reinforcement Learning},
  journal      = {CoRR},
  volume       = {abs/2004.07219},
  year         = {2020},
  url          = {https://arxiv.org/abs/2004.07219},
  eprinttype    = {arXiv},
  eprint       = {2004.07219},
  bibsource    = {dblp computer science bibliography, https://dblp.org}
}

@inproceedings{DBLP:conf/nips/KumarZTL20,
  author       = {Aviral Kumar and
                  Aurick Zhou and
                  George Tucker and
                  Sergey Levine},
  editor       = {Hugo Larochelle and
                  Marc'Aurelio Ranzato and
                  Raia Hadsell and
                  Maria{-}Florina Balcan and
                  Hsuan{-}Tien Lin},
  title        = {Conservative Q-Learning for Offline Reinforcement Learning},
  booktitle    = {Advances in Neural Information Processing Systems 33: Annual Conference
                  on Neural Information Processing Systems 2020, NeurIPS 2020, December
                  6-12, 2020, virtual},
  year         = {2020},
  url          = {https://proceedings.neurips.cc/paper/2020/hash/0d2b2061826a5df3221116a5085a6052-Abstract.html},
  bibsource    = {dblp computer science bibliography, https://dblp.org}
}

@inproceedings{DBLP:conf/iclr/KostrikovNL22,
  author       = {Ilya Kostrikov and
                  Ashvin Nair and
                  Sergey Levine},
  title        = {Offline Reinforcement Learning with Implicit Q-Learning},
  booktitle    = {The Tenth International Conference on Learning Representations, {ICLR}
                  2022, Virtual Event, April 25-29, 2022},
  publisher    = {OpenReview.net},
  year         = {2022},
  url          = {https://openreview.net/forum?id=68n2s9ZJWF8},
  bibsource    = {dblp computer science bibliography, https://dblp.org}
}

@inproceedings{DBLP:conf/nips/RigterLH23,
  author       = {Marc Rigter and
                  Bruno Lacerda and
                  Nick Hawes},
  editor       = {Alice Oh and
                  Tristan Naumann and
                  Amir Globerson and
                  Kate Saenko and
                  Moritz Hardt and
                  Sergey Levine},
  title        = {One Risk to Rule Them All: {A} Risk-Sensitive Perspective on Model-Based
                  Offline Reinforcement Learning},
  booktitle    = {Advances in Neural Information Processing Systems 36: Annual Conference
                  on Neural Information Processing Systems 2023, NeurIPS 2023, New Orleans,
                  LA, USA, December 10 - 16, 2023},
  year         = {2023},
  url          = {http://papers.nips.cc/paper\_files/paper/2023/hash/f49287371916715b9209fa41a275851e-Abstract-Conference.html},
  bibsource    = {dblp computer science bibliography, https://dblp.org}
}

@inproceedings{DBLP:conf/nips/JannerLL21,
  author       = {Michael Janner and
                  Qiyang Li and
                  Sergey Levine},
  editor       = {Marc'Aurelio Ranzato and
                  Alina Beygelzimer and
                  Yann N. Dauphin and
                  Percy Liang and
                  Jennifer Wortman Vaughan},
  title        = {Offline Reinforcement Learning as One Big Sequence Modeling Problem},
  booktitle    = {Advances in Neural Information Processing Systems 34: Annual Conference
                  on Neural Information Processing Systems 2021, NeurIPS 2021, December
                  6-14, 2021, virtual},
  pages        = {1273--1286},
  year         = {2021},
  url          = {https://proceedings.neurips.cc/paper/2021/hash/099fe6b0b444c23836c4a5d07346082b-Abstract.html},
  bibsource    = {dblp computer science bibliography, https://dblp.org}
}

@inproceedings{DBLP:conf/iclr/JiangZJLRGT23,
  author       = {Zhengyao Jiang and
                  Tianjun Zhang and
                  Michael Janner and
                  Yueying Li and
                  Tim Rockt{\"{a}}schel and
                  Edward Grefenstette and
                  Yuandong Tian},
  title        = {Efficient Planning in a Compact Latent Action Space},
  booktitle    = {The Eleventh International Conference on Learning Representations,
                  {ICLR} 2023, Kigali, Rwanda, May 1-5, 2023},
  publisher    = {OpenReview.net},
  year         = {2023},
  url          = {https://openreview.net/forum?id=cA77NrVEuqn},
  bibsource    = {dblp computer science bibliography, https://dblp.org}
}

@inproceedings{DBLP:conf/iclr/AntonoglouSOHS22,
  author       = {Ioannis Antonoglou and
                  Julian Schrittwieser and
                  Sherjil Ozair and
                  Thomas K. Hubert and
                  David Silver},
  title        = {Planning in Stochastic Environments with a Learned Model},
  booktitle    = {The Tenth International Conference on Learning Representations, {ICLR}
                  2022, Virtual Event, April 25-29, 2022},
  publisher    = {OpenReview.net},
  year         = {2022},
  url          = {https://openreview.net/forum?id=X6D9bAHhBQ1},
  bibsource    = {dblp computer science bibliography, https://dblp.org}
}

@inproceedings{DBLP:conf/nips/OordVK17,
  author       = {A{\"{a}}ron van den Oord and
                  Oriol Vinyals and
                  Koray Kavukcuoglu},
  editor       = {Isabelle Guyon and
                  Ulrike von Luxburg and
                  Samy Bengio and
                  Hanna M. Wallach and
                  Rob Fergus and
                  S. V. N. Vishwanathan and
                  Roman Garnett},
  title        = {Neural Discrete Representation Learning},
  booktitle    = {Advances in Neural Information Processing Systems 30: Annual Conference
                  on Neural Information Processing Systems 2017, December 4-9, 2017,
                  Long Beach, CA, {USA}},
  pages        = {6306--6315},
  year         = {2017},
  url          = {https://proceedings.neurips.cc/paper/2017/hash/7a98af17e63a0ac09ce2e96d03992fbc-Abstract.html},
  bibsource    = {dblp computer science bibliography, https://dblp.org}
}

@inproceedings{DBLP:conf/icml/HubertSABSS21,
  author       = {Thomas Hubert and
                  Julian Schrittwieser and
                  Ioannis Antonoglou and
                  Mohammadamin Barekatain and
                  Simon Schmitt and
                  David Silver},
  editor       = {Marina Meila and
                  Tong Zhang},
  title        = {Learning and Planning in Complex Action Spaces},
  booktitle    = {Proceedings of the 38th International Conference on Machine Learning,
                  {ICML} 2021, 18-24 July 2021, Virtual Event},
  series       = {Proceedings of Machine Learning Research},
  volume       = {139},
  pages        = {4476--4486},
  publisher    = {{PMLR}},
  year         = {2021},
  url          = {http://proceedings.mlr.press/v139/hubert21a.html},
  bibsource    = {dblp computer science bibliography, https://dblp.org}
}

@misc{silver2017masteringchessshogiselfplay,
      title={Mastering Chess and Shogi by Self-Play with a General Reinforcement Learning Algorithm}, 
      author={David Silver and Thomas Hubert and Julian Schrittwieser and Ioannis Antonoglou and Matthew Lai and Arthur Guez and Marc Lanctot and Laurent Sifre and Dharshan Kumaran and Thore Graepel and Timothy Lillicrap and Karen Simonyan and Demis Hassabis},
      year={2017},
      eprint={1712.01815},
      archivePrefix={arXiv},
      primaryClass={cs.AI},
      url={https://arxiv.org/abs/1712.01815}, 
}

@article{DBLP:journals/nature/SchrittwieserAH20,
  author       = {Julian Schrittwieser and
                  Ioannis Antonoglou and
                  Thomas Hubert and
                  Karen Simonyan and
                  Laurent Sifre and
                  Simon Schmitt and
                  Arthur Guez and
                  Edward Lockhart and
                  Demis Hassabis and
                  Thore Graepel and
                  Timothy P. Lillicrap and
                  David Silver},
  title        = {Mastering Atari, Go, chess and shogi by planning with a learned model},
  journal      = {Nat.},
  volume       = {588},
  number       = {7839},
  pages        = {604--609},
  year         = {2020},
  url          = {https://doi.org/10.1038/s41586-020-03051-4},
  doi          = {10.1038/S41586-020-03051-4},
  bibsource    = {dblp computer science bibliography, https://dblp.org}
}

@inproceedings{DBLP:conf/corl/LuoDWKGL23,
  author       = {Jianlan Luo and
                  Perry Dong and
                  Jeffrey Wu and
                  Aviral Kumar and
                  Xinyang Geng and
                  Sergey Levine},
  editor       = {Jie Tan and
                  Marc Toussaint and
                  Kourosh Darvish},
  title        = {Action-Quantized Offline Reinforcement Learning for Robotic Skill
                  Learning},
  booktitle    = {Conference on Robot Learning, CoRL 2023, 6-9 November 2023, Atlanta,
                  GA, {USA}},
  series       = {Proceedings of Machine Learning Research},
  volume       = {229},
  pages        = {1348--1361},
  publisher    = {{PMLR}},
  year         = {2023},
  url          = {https://proceedings.mlr.press/v229/luo23a.html},
  bibsource    = {dblp computer science bibliography, https://dblp.org}
}

@inproceedings{DBLP:conf/nips/ChenLRLGLASM21,
  author       = {Lili Chen and
                  Kevin Lu and
                  Aravind Rajeswaran and
                  Kimin Lee and
                  Aditya Grover and
                  Michael Laskin and
                  Pieter Abbeel and
                  Aravind Srinivas and
                  Igor Mordatch},
  editor       = {Marc'Aurelio Ranzato and
                  Alina Beygelzimer and
                  Yann N. Dauphin and
                  Percy Liang and
                  Jennifer Wortman Vaughan},
  title        = {Decision Transformer: Reinforcement Learning via Sequence Modeling},
  booktitle    = {Advances in Neural Information Processing Systems 34: Annual Conference
                  on Neural Information Processing Systems 2021, NeurIPS 2021, December
                  6-14, 2021, virtual},
  pages        = {15084--15097},
  year         = {2021},
  url          = {https://proceedings.neurips.cc/paper/2021/hash/7f489f642a0ddb10272b5c31057f0663-Abstract.html},
  bibsource    = {dblp computer science bibliography, https://dblp.org}
}

@inproceedings{DBLP:conf/nips/PasterMB22,
  author       = {Keiran Paster and
                  Sheila A. McIlraith and
                  Jimmy Ba},
  editor       = {Sanmi Koyejo and
                  S. Mohamed and
                  A. Agarwal and
                  Danielle Belgrave and
                  K. Cho and
                  A. Oh},
  title        = {You Can't Count on Luck: Why Decision Transformers and RvS Fail in
                  Stochastic Environments},
  booktitle    = {Advances in Neural Information Processing Systems 35: Annual Conference
                  on Neural Information Processing Systems 2022, NeurIPS 2022, New Orleans,
                  LA, USA, November 28 - December 9, 2022},
  year         = {2022},
  url          = {http://papers.nips.cc/paper\_files/paper/2022/hash/fe90657b12193c7b52a3418bdc351807-Abstract-Conference.html},
  bibsource    = {dblp computer science bibliography, https://dblp.org}
}

@inproceedings{DBLP:conf/nips/VaswaniSPUJGKP17,
  author       = {Ashish Vaswani and
                  Noam Shazeer and
                  Niki Parmar and
                  Jakob Uszkoreit and
                  Llion Jones and
                  Aidan N. Gomez and
                  Lukasz Kaiser and
                  Illia Polosukhin},
  editor       = {Isabelle Guyon and
                  Ulrike von Luxburg and
                  Samy Bengio and
                  Hanna M. Wallach and
                  Rob Fergus and
                  S. V. N. Vishwanathan and
                  Roman Garnett},
  title        = {Attention is All you Need},
  booktitle    = {Advances in Neural Information Processing Systems 30: Annual Conference
                  on Neural Information Processing Systems 2017, December 4-9, 2017,
                  Long Beach, CA, {USA}},
  pages        = {5998--6008},
  year         = {2017},
  url          = {https://proceedings.neurips.cc/paper/2017/hash/3f5ee243547dee91fbd053c1c4a845aa-Abstract.html},
  bibsource    = {dblp computer science bibliography, https://dblp.org}
}

@inproceedings{
luo2025scalable,
title={Scalable Decision-Making in Stochastic Environments through Learned Temporal Abstraction},
author={Baiting Luo and Ava Pettet and Aron Laszka and Abhishek Dubey and Ayan Mukhopadhyay},
booktitle={The Thirteenth International Conference on Learning Representations},
year={2025},
url={https://openreview.net/forum?id=pQsllTesiE}
}

@article{
doi:10.1126/science.aar6404,
author = {David Silver  and Thomas Hubert  and Julian Schrittwieser  and Ioannis Antonoglou  and Matthew Lai  and Arthur Guez  and Marc Lanctot  and Laurent Sifre  and Dharshan Kumaran  and Thore Graepel  and Timothy Lillicrap  and Karen Simonyan  and Demis Hassabis },
title = {A general reinforcement learning algorithm that masters chess, shogi, and Go through self-play},
journal = {Science},
volume = {362},
number = {6419},
pages = {1140-1144},
year = {2018},
doi = {10.1126/science.aar6404},
URL = {https://www.science.org/doi/abs/10.1126/science.aar6404},
eprint = {https://www.science.org/doi/pdf/10.1126/science.aar6404}}

@inproceedings{wind,
    author = {Popko, Wojciech and Wächter, Matthias and Thomas, Philipp},
    title = {Verification of Continuous Time Random Walk Wind Model},
    volume = {All Days},
    series = {International Ocean and Polar Engineering Conference},
    pages = {ISOPE-I-16-189},
    year = {2016},
    month = {06},
}

@inproceedings{
son2025distilling,
title={Distilling Reinforcement Learning Algorithms for In-Context Model-Based Planning},
author={Jaehyeon Son and Soochan Lee and Gunhee Kim},
booktitle={The Thirteenth International Conference on Learning Representations},
year={2025},
url={https://openreview.net/forum?id=BfUugGfBE5}
}

@inproceedings{DBLP:conf/icml/LiuA23,
  author       = {Hao Liu and
                  Pieter Abbeel},
  editor       = {Andreas Krause and
                  Emma Brunskill and
                  Kyunghyun Cho and
                  Barbara Engelhardt and
                  Sivan Sabato and
                  Jonathan Scarlett},
  title        = {Emergent Agentic Transformer from Chain of Hindsight Experience},
  booktitle    = {International Conference on Machine Learning, {ICML} 2023, 23-29 July
                  2023, Honolulu, Hawaii, {USA}},
  series       = {Proceedings of Machine Learning Research},
  volume       = {202},
  pages        = {21362--21374},
  publisher    = {{PMLR}},
  year         = {2023},
  url          = {https://proceedings.mlr.press/v202/liu23a.html},
  bibsource    = {dblp computer science bibliography, https://dblp.org}
}

@inproceedings{DBLP:conf/iclr/LaskinWOPSSSHFB23,
  author       = {Michael Laskin and
                  Luyu Wang and
                  Junhyuk Oh and
                  Emilio Parisotto and
                  Stephen Spencer and
                  Richie Steigerwald and
                  DJ Strouse and
                  Steven Stenberg Hansen and
                  Angelos Filos and
                  Ethan Brooks and
                  Maxime Gazeau and
                  Himanshu Sahni and
                  Satinder Singh and
                  Volodymyr Mnih},
  title        = {In-context Reinforcement Learning with Algorithm Distillation},
  booktitle    = {The Eleventh International Conference on Learning Representations,
                  {ICLR} 2023, Kigali, Rwanda, May 1-5, 2023},
  publisher    = {OpenReview.net},
  year         = {2023},
  url          = {https://openreview.net/forum?id=hy0a5MMPUv},
  bibsource    = {dblp computer science bibliography, https://dblp.org}
}

@inproceedings{DBLP:conf/nips/0002XPCFNB23,
  author       = {Jonathan Lee and
                  Annie Xie and
                  Aldo Pacchiano and
                  Yash Chandak and
                  Chelsea Finn and
                  Ofir Nachum and
                  Emma Brunskill},
  editor       = {Alice Oh and
                  Tristan Naumann and
                  Amir Globerson and
                  Kate Saenko and
                  Moritz Hardt and
                  Sergey Levine},
  title        = {Supervised Pretraining Can Learn In-Context Reinforcement Learning},
  booktitle    = {Advances in Neural Information Processing Systems 36: Annual Conference
                  on Neural Information Processing Systems 2023, NeurIPS 2023, New Orleans,
                  LA, USA, December 10 - 16, 2023},
  year         = {2023},
  url          = {http://papers.nips.cc/paper\_files/paper/2023/hash/8644b61a9bc87bf7844750a015feb600-Abstract-Conference.html},
  bibsource    = {dblp computer science bibliography, https://dblp.org}
}

@inproceedings{
furuta2022generalized,
title={Generalized Decision Transformer for Offline Hindsight Information Matching},
author={Hiroki Furuta and Yutaka Matsuo and Shixiang Shane Gu},
booktitle={International Conference on Learning Representations},
year={2022},
url={https://openreview.net/forum?id=CAjxVodl_v}
}

@inproceedings{DBLP:conf/icml/HuangHCS024,
  author       = {Sili Huang and
                  Jifeng Hu and
                  Hechang Chen and
                  Lichao Sun and
                  Bo Yang},
  title        = {In-Context Decision Transformer: Reinforcement Learning via Hierarchical
                  Chain-of-Thought},
  booktitle    = {Forty-first International Conference on Machine Learning, {ICML} 2024,
                  Vienna, Austria, July 21-27, 2024},
  publisher    = {OpenReview.net},
  year         = {2024},
  url          = {https://openreview.net/forum?id=jmmji1EU3g},
  bibsource    = {dblp computer science bibliography, https://dblp.org}
}

@article{macro-action,
author = {Mcgovern, Amy and Sutton, Richard},
year = {1998},
month = {10},
pages = {},
title = {Macro-Actions in Reinforcement Learning: An Empirical Analysis}
}

@inproceedings{DBLP:conf/iclr/JiangXWLJGRT24,
  author       = {Zhengyao Jiang and
                  Yingchen Xu and
                  Nolan Wagener and
                  Yicheng Luo and
                  Michael Janner and
                  Edward Grefenstette and
                  Tim Rockt{\"{a}}schel and
                  Yuandong Tian},
  title        = {{H-GAP:} Humanoid Control with a Generalist Planner},
  booktitle    = {The Twelfth International Conference on Learning Representations,
                  {ICLR} 2024, Vienna, Austria, May 7-11, 2024},
  publisher    = {OpenReview.net},
  year         = {2024},
  url          = {https://openreview.net/forum?id=LYG6tBlEX0},
  bibsource    = {dblp computer science bibliography, https://dblp.org}
}

@inproceedings{DBLP:conf/nips/BrownMRSKDNSSAA20,
  author       = {Tom B. Brown and
                  Benjamin Mann and
                  Nick Ryder and
                  Melanie Subbiah and
                  Jared Kaplan and
                  Prafulla Dhariwal and
                  Arvind Neelakantan and
                  Pranav Shyam and
                  Girish Sastry and
                  Amanda Askell and
                  Sandhini Agarwal and
                  Ariel Herbert{-}Voss and
                  Gretchen Krueger and
                  Tom Henighan and
                  Rewon Child and
                  Aditya Ramesh and
                  Daniel M. Ziegler and
                  Jeffrey Wu and
                  Clemens Winter and
                  Christopher Hesse and
                  Mark Chen and
                  Eric Sigler and
                  Mateusz Litwin and
                  Scott Gray and
                  Benjamin Chess and
                  Jack Clark and
                  Christopher Berner and
                  Sam McCandlish and
                  Alec Radford and
                  Ilya Sutskever and
                  Dario Amodei},
  editor       = {Hugo Larochelle and
                  Marc'Aurelio Ranzato and
                  Raia Hadsell and
                  Maria{-}Florina Balcan and
                  Hsuan{-}Tien Lin},
  title        = {Language Models are Few-Shot Learners},
  booktitle    = {Advances in Neural Information Processing Systems 33: Annual Conference
                  on Neural Information Processing Systems 2020, NeurIPS 2020, December
                  6-12, 2020, virtual},
  year         = {2020},
  url          = {https://proceedings.neurips.cc/paper/2020/hash/1457c0d6bfcb4967418bfb8ac142f64a-Abstract.html},
  bibsource    = {dblp computer science bibliography, https://dblp.org}
}

@inproceedings{DBLP:conf/cvpr/LeeKKCH22,
  author       = {Doyup Lee and
                  Chiheon Kim and
                  Saehoon Kim and
                  Minsu Cho and
                  Wook{-}Shin Han},
  title        = {Autoregressive Image Generation using Residual Quantization},
  booktitle    = {{IEEE/CVF} Conference on Computer Vision and Pattern Recognition,
                  {CVPR} 2022, New Orleans, LA, USA, June 18-24, 2022},
  pages        = {11513--11522},
  publisher    = {{IEEE}},
  year         = {2022},
  url          = {https://doi.org/10.1109/CVPR52688.2022.01123},
  doi          = {10.1109/CVPR52688.2022.01123},
  bibsource    = {dblp computer science bibliography, https://dblp.org}
}

@inproceedings{DBLP:conf/rss/RajeswaranKGVST18,
  author       = {Aravind Rajeswaran and
                  Vikash Kumar and
                  Abhishek Gupta and
                  Giulia Vezzani and
                  John Schulman and
                  Emanuel Todorov and
                  Sergey Levine},
  editor       = {Hadas Kress{-}Gazit and
                  Siddhartha S. Srinivasa and
                  Tom Howard and
                  Nikolay Atanasov},
  title        = {Learning Complex Dexterous Manipulation with Deep Reinforcement Learning
                  and Demonstrations},
  booktitle    = {Robotics: Science and Systems XIV, Carnegie Mellon University, Pittsburgh,
                  Pennsylvania, USA, June 26-30, 2018},
  year         = {2018},
  url          = {http://www.roboticsproceedings.org/rss14/p49.html},
  doi          = {10.15607/RSS.2018.XIV.049},
  bibsource    = {dblp computer science bibliography, https://dblp.org}
}
\bibliographystyle{unsrtnat}
\appendix
\section{Appendix}
\subsection{Latent Search Tree Construction}\label{sec:algo}

\begin{breakablealgorithm}
\small
\caption{\textsc{SampleStack}$(p_{\phi}, \mathcal{C}, D, \boldsymbol{\xi}, \boldsymbol{\rho})$}
\label{alg:samplestack_macro}
\begin{algorithmic}[1]
  \State $\mathbf{u}_{1} \gets \text{scores from } p_{\phi}(k_{1}\mid \mathcal{C})$
  \State $k_{1} \sim \TopKTempCat\!\big(\mathbf{u}_{1},\,\xi_1,\,\rho_1\big)$
  \State $K \gets (k_{1})$
  \For{$d=2$ \textbf{to} $D$}
    \State $\mathbf{u}_{d} \gets \text{scores from } p_{\phi}(k_{d}\mid K_{1:d-1},\,\mathcal{C})$
    \State $k_{d} \sim \TopKTempCat\!\big(\mathbf{u}_{d},\,\xi_d,\,\rho_d\big)$
    \State $K \gets (K,\,k_{d})$ \Comment{append}
  \EndFor
  \State \Return $K$
\end{algorithmic}
\end{breakablealgorithm}

\begin{breakablealgorithm}
\small
\caption{\textsc{Pre-constructing the Latent Search Space (Residual Stack, Macro Tokens)}}
\label{alg:prebuild_latent_space_residual_macro}
\begin{algorithmic}[1]
  \Require Current observation $o_{t_k}$; context $\bigl((G^{(L)}_{t-cL},o_{t-cL}, a_{t-cL}),\dots,(G^{(L)}_{t-1}, o_{t-1}, a_{t-1})\bigr)$; 
           encoder $f_{\mathrm{enc}}$; decoder $f_{\mathrm{dec}}$; residual-stack model $p_{\phi}$;
           residual depth $D$; per-depth temperatures $\boldsymbol{\xi}=(\xi_1,\ldots,\xi_D)$;
           per-depth top truncation $\boldsymbol{\rho}=(\rho_1,\ldots,\rho_D)$;
           \# coarse samples $M$; \# residual completions per coarse sample $J$;
           \# lookahead samples $N$; tree depth $H$; \# kept per node $\kappa_{\mathrm{keep}}$;
           \# proposals per node $B$
  \Ensure Latent search tree $\mathcal{T}$ with cached promising residual-stack codes

  \State \textbf{Encode macro-context}
  \State $k_{t_{k-1}:t_{k-c},\,1:D} \gets f_{\mathrm{enc}}\!\bigl((G^{(L)}_{t-cL},o_{t-cL}, a_{t-cL}),\dots,(G^{(L)}_{t-1}, o_{t-1}, a_{t-1})\bigr)$
  \State Initialize tree $\mathcal{T}$ with root node $s_{t_k}=(o_{t_k},\,\pastK)$
  \State $\Ck \gets (o_{t_k},\,\pastK)$

  \State \textbf{Step 1: sample and score initial macro stacks at index $k$}
  \State \textit{/* $M$ coarse draws for depth 1; for each, $J$ residual completions to depth $D$ */}
  \State $\mathbf{u}_{k,1} \gets \text{scores from } p_{\phi}\!\big(k_{t_k,1}\mid \Ck\big)$
  \For{$i=1$ \textbf{to} $M$ \textbf{(parallel)}}
    \State $k_{t_k,1}^{(i)} \sim \TopKTempCat\!\big(\mathbf{u}_{k,1},\,\xi_1,\,\rho_1\big)$
    \For{$j=1$ \textbf{to} $J$ \textbf{(parallel)}}
      \State $K_{t_k}^{(i,j)} \gets (k_{t_k,1}^{(i)})$
      \For{$d=2$ \textbf{to} $D$}
        \State $\mathbf{u}_{k,d} \gets \text{scores from } p_{\phi}\!\big(k_{t_k,d}\mid K_{t_k,1:d-1}^{(i,j)},\,\Ck\big)$
        \State $k_{t_k,d}^{(i,j)} \sim \TopKTempCat\!\big(\mathbf{u}_{k,d},\,\xi_d,\,\rho_d\big)$
        \State $K_{t_k}^{(i,j)} \gets (K_{t_k}^{(i,j)},\,k_{t_k,d}^{(i,j)})$ \Comment{append}
      \EndFor
      \State $z_{t_k}^{(i,j)} \gets \Emb{K_{t_k}^{(i,j)}}$
      \State $\widehat{G}^{(L)}_{t_k}\!\big(K_{t_k}^{(i,j)}\big) \gets \text{current-step head from } f_{\mathrm{dec}}$
      \For{$n=1$ \textbf{to} $N$ \textbf{(parallel)}}
         \State $\mathcal{C}' \gets (o_{t_k},\,K_{t_k}^{(i,j)},\,\pastK)$
         \State $K_{t_{k+1}}^{(i,j,n)} \gets \SampleStack(p_{\phi},\,\mathcal{C}',\,D,\,\boldsymbol{\xi},\,\boldsymbol{\rho})$
         \State $z_{t_{k+1}}^{(i,j,n)} \gets \Emb{K_{t_{k+1}}^{(i,j,n)}}$
         \State $\hat{y}_{t_{k+1}}^{(i,j,n)} \gets
                f_{\mathrm{dec}}\!\bigl(z_{t_k}^{(i,j)},\,z_{t_{k+1}}^{(i,j,n)},\,o_{t_k},\,\pastK\bigr)$
      \EndFor
      \State $\text{score}\!\big(K_{t_k}^{(i,j)}\big) \gets 
             \frac{1}{N}\sum_{n=1}^{N}\Big(\widehat{G}^{(L)}_{t_k}\!\big(K_{t_k}^{(i,j)}\big) + \RTG{\hat{y}_{t_{k+1}}^{(i,j,n)}}\Big)$
      \State $\bar{y}_{t_{k+1}}^{(i,j)} \gets \frac{1}{N}\sum_{n=1}^{N}\hat{y}_{t_{k+1}}^{(i,j,n)}$;\quad
             $\hat{o}_{t_{k+1}}^{(i,j)} \gets \Obs\!\big(\bar{y}_{t_{k+1}}^{(i,j)}\big)$
    \EndFor
  \EndFor
  \State Select top-$\kappa_{\mathrm{keep}}$ stacks $\{K_{t_k}^{(i,j)}\}$ by $\text{score}$; 
         for each, attach child node $\big(\hat{o}_{t_{k+1}}^{(i,j)},\,K_{t_k}^{(i,j)}\big)$ under the root in $\mathcal{T}$

  \State \textbf{Step 2: recursive latent-tree expansion over macro indices}
  \For{$h=2$ \textbf{to} $H$}
    \State Let $\mathcal{N}_{h-1}$ be the nodes at depth $h-1$ of $\mathcal{T}$
    \For{\textbf{each} node $(\hat o, K^{c}) \in \mathcal{N}_{h-1}$ \textbf{(parallel)}}
      \State $\mathcal{C} \gets (\hat o,\,K^{c},\,\pastK)$
      \For{$b=1$ \textbf{to} $B$ \textbf{(parallel)}}
        \State $K_{t_{k+h-1}}^{(b)} \gets \SampleStack(p_{\phi},\,\mathcal{C},\,D,\,\boldsymbol{\xi},\,\boldsymbol{\rho})$
        \State $z_{t_{k+h-1}}^{(b)} \gets \Emb{K_{t_{k+h-1}}^{(b)}}$
        \State $\widehat{G}^{(L)}_{t_{k+h-1}}\!\big(K_{t_{k+h-1}}^{(b)}\big) \gets \text{current-step head from } f_{\mathrm{dec}}$
        \For{$n=1$ \textbf{to} $N$ \textbf{(parallel)}}
          \State $\mathcal{C}^\star \gets (\hat o,\,K_{t_{k+h-1}}^{(b)},\,\pastK)$
          \State $K_{t_{k+h}}^{(b,n)} \gets \SampleStack(p_{\phi},\,\mathcal{C}^\star,\,D,\,\boldsymbol{\xi},\,\boldsymbol{\rho})$
          \State $z_{t_{k+h}}^{(b,n)} \gets \Emb{K_{t_{k+h}}^{(b,n)}}$
          \State $\hat{y}_{t_{k+h}}^{(b,n)} \gets
                 f_{\mathrm{dec}}\!\bigl(z_{t_{k+h-1}}^{(b)},\,z_{t_{k+h}}^{(b,n)},\,\hat o,\,\pastK\bigr)$
        \EndFor
        \State $\text{score}\!\big(K_{t_{k+h-1}}^{(b)}\big) \gets 
               \frac{1}{N}\sum_{n=1}^{N}\Big(\widehat{G}^{(L)}_{t_{k+h-1}}\!\big(K_{t_{k+h-1}}^{(b)}\big) + \RTG{\hat{y}_{t_{k+h}}^{(b,n)}}\Big)$
        \State $\bar{y}_{t_{k+h}}^{(b)} \gets \frac{1}{N}\sum_{n=1}^{N}\hat{y}_{t_{k+h}}^{(b,n)}$;\quad
               $\hat{o}_{t_{k+h}}^{(b)} \gets \Obs\!\big(\bar{y}_{t_{k+h}}^{(b)}\big)$
      \EndFor
      \State Select top-$\kappa_{\mathrm{keep}}$ from $\{K_{t_{k+h-1}}^{(b)}\}_{b=1}^B$ by $\text{score}$ and attach as children 
             $(\hat{o}_{t_{k+h}}^{(b)},\,K_{t_{k+h-1}}^{(b)})$ under $(\hat o, K^{c})$ in $\mathcal{T}$
    \EndFor
  \EndFor

  \State \Return $\mathcal{T}$
\end{algorithmic}
\end{breakablealgorithm}

\subsection{Experiment Details}\label{sec:experiment_domains}
\subsubsection{Implementation Details}\label{sec:parameters}
The hyperparameter settings for \itap{} are listed in Table~\ref{tab:hyperparams}.
For baselines, we follow the implementations and recommended hyperparameters from the original works, including
TAP~\citep{DBLP:conf/iclr/JiangZJLRGT23}, L-MAP~\citep{luo2025scalable}, CQL~\citep{DBLP:conf/nips/KumarZTL20},
IQL~\cite{DBLP:conf/iclr/KostrikovNL22}, DT~\citep{DBLP:conf/nips/ChenLRLGLASM21}, and 1R2R~\citep{DBLP:conf/nips/RigterLH23}. For the Stochastic MuJoCo benchmarks, the results of TAP, 1R2R, CQL, and IQL are taken from the L-MAP paper~\citep{luo2025scalable},
which evaluates these methods under the same stochastic-domain protocol.
For deterministic domains, the CQL and IQL results are taken from their respective papers~\citep{DBLP:conf/nips/KumarZTL20,DBLP:conf/iclr/KostrikovNL22}.
Each run of \itap{} takes approximately 6 hours on one NVIDIA RTX 5090 GPU with an Intel(R) Core(TM) i9-14900KS CPU.


\begin{table}[htbp]
  \centering
  \caption{List of Hyper-parameters}
  \label{tab:hyperparams}
  \begin{tabular}{lll}
    \toprule
    \textbf{Environment} & \textbf{Hyper-parameter}                     & \textbf{Value} \\ \midrule
    All                  & learning rate                                & $1\times10^{-4}$ \\
    All                  & batch size                                   & 512 \\
    All                  & dropout probability                          & 0.1 \\
    All                  & number of attention heads                    & 4 \\
    All                  & macro action length $L$                      & 3 \\
    All                  & embedding size (latent code)                 & 512 \\
    All                  & $c_1$                                        & 1.25 \\
    All                  & $c_2$                                        & 19652\\
    All                  & $\alpha_{\text{tail}}$                       & 1\\
    All                  & $\alpha_{\text{ctx}}$                        & 0.1\\
    All                  & $\beta_{\mathrm{ps}}$                        & 1\\
    \midrule
    MuJoco   & context length $c$$/$training sequence length  & 6/24 \\
    MuJoco   & discount factor                              & 0.99 \\
    MuJoco   & number of Transformer layers                 & 4 \\
    MuJoco   & feature vector size                          & 512 \\
    MuJoco   & codebook size                          & 512 \\
    MuJoco               & initial number of policy samples $M$                   & 16 \\
    MuJoco               & number of transition samples $N$                   & 4 \\
    MuJoco               & number of policy samples $B$                   & 4 \\
    MuJoco   & number of MCTS iterations                             & 100 \\
    Mujoco   & $\kappa_{\mathrm{keep}}$ & $50\%$ \\
    MuJoco   & temperature                             & 2 \\
    MuJoco   & Residual Depth                             & 1 \\
    MuJoco   & primitive planning horizon                             & 9 \\ \midrule
    Adroit               & context length $c$$/$training sequence length & 6$/$24 \\
    Adroit               & discount factor                              & 0.99 \\
    Adroit               & number Transformer layers                    & 4 \\
    Adroit               & feature vector size                          & 256 \\
    Adroit               & codebook size                                & 512 \\
    Adroit               & initial number of policy samples $M$                   & 16 \\
    Adroit               & number of transition samples $N$                   & 4 \\
    Adroit               & number of policy samples $B$                   & 4 \\
    Adroit               & number of MCTS iterations                   & 100 \\
    Adroit               & Residual Depth                    & 2,3 \\
    Adroit               & $\kappa_{\mathrm{keep}}$   & $10\%$ \\
    Adroit               & $J$   & 4 \\
    Adroit               & temperature   & 1 \\
    Adroit               & primitive planning horizon                             & 9 \\ 
    \bottomrule
  \end{tabular}
\end{table}

\subsubsection{Domains}\label{appendix:domain}


\paragraph{Stochastic MuJoCo.}
The Stochastic MuJoCo tasks introduced by \citet{DBLP:conf/nips/RigterLH23} apply incremental perturbation forces following a uniform random walk~\citep{wind}. At each timestep, the perturbation force \( f_t \) is updated as:
\begin{equation}
    f_{t+1} = f_t + \Delta f, \quad \Delta f \sim \text{Uniform}\left(-0.1 \cdot f_{\text{MAX}},\; 0.1 \cdot f_{\text{MAX}}\right),
\end{equation}
with the total perturbation clipped to remain within \([-f_{\text{MAX}}, f_{\text{MAX}}]\). This model introduces persistent, incremental perturbations, representing a baseline scenario. For both Hopper and Walker2D environments, the perturbation force \(f_t\) is applied horizontally along the x-axis to simulate external disturbances, such as wind gusts. The maximum perturbation magnitude \(f_{\text{MAX}}\) specifically for the Hopper moderate perturbation level is 2.5 Newtons, high perturbation level is 5 Newtons, Walker2D moderate perturbation level is 7 Newtons, high perturbation level is 12 Newtons.

\paragraph{Adroit as a POMDP.}
Many Adroit manipulation tasks expose \emph{privileged} channels that directly encode the goal state and progress (e.g., target poses, object--target deltas, insertion depth). To evaluate agents under partial observability, we define a POMDP variant that preserves the original dynamics and reward but \emph{ablates} these channels at evaluation time. Concretely, let the environment produce an observation $o_t \in \mathbb{R}^d$ under the standard \texttt{v1} layout. We introduce a fixed binary mask $m \in \{0,1\}^d$ (zeros at privileged indices) and set
\[
\tilde{o}_t \;=\; m \odot o_t, \qquad 
O(\tilde{o}_t \mid s_t) \;=\; \delta\!\big(\tilde{o}_t - m \odot o_t\big),
\]
yielding a POMDP in which progress toward the goal must be \emph{inferred} from history rather than read off directly.

\textbf{Masking regimes.} Indices refer to 0\,-based positions in the default \texttt{v1} observation vector.
\begin{itemize}
\item \textbf{Pen} ($d{=}45$): $\mathcal{I}=\{36,37,39,40,42,43\}$.
\item \textbf{Relocate} ($d{=}39$): $\mathcal{I}=\{30,\ldots,38\}$.
\item \textbf{Door} ($d{=}39$): $\mathcal{I}=\{27,28,32,\ldots,38\}$.
\item \textbf{Hammer} ($d{=}46$): $\mathcal{I}=\{43,44,45\}$.
\end{itemize}

This construction leaves proprioception and contact signals intact while removing privileged goal vectors and progress proxies, thereby converting the original fully observable tasks into history-dependent POMDPs that better reflect realistic sensing and require temporal credit assignment and state estimation.


\end{document}